\documentclass[preprint,12pt,authoryear,a4paper]{elsarticle}
\usepackage[T1]{fontenc}
\usepackage{lmodern}
\usepackage{graphicx}
\usepackage{booktabs}
\usepackage[table]{xcolor}
\usepackage{placeins}
\usepackage{amsmath}
\usepackage{amssymb}
\usepackage{longtable}
\usepackage{multirow}
\usepackage{hyperref}
\hypersetup{hidelinks}
\bibpunct{(}{)}{;}{a}{,}{,}
\biboptions{sort}
\newcommand{\nc}[1]{\textcolor{black!45}{#1}}
\newcommand{\gooddelta}[1]{\textcolor{green!50!black}{#1}}
\newcommand{\baddelta}[1]{\textcolor{red!70!black}{#1}}
\newcommand{\ROMANAuthor}{Gonzalo Uribarri}
\newcommand{\ROMANEmail}{gonzalo.uribarri@dsv.su.se}
\newcommand{\ROMANAddressLine}{Nodhuset, Borgarfjordsgatan 12}
\newcommand{\ROMANPostcode}{164 25}
\newcommand{\ROMANAuthorBlock}{%
  \author[dsv]{\ROMANAuthor\texorpdfstring{\corref{cor1}}{}}
  \if\relax\detokenize\expandafter{\ROMANEmail}\relax\else\ead{\ROMANEmail}\fi
  \cortext[cor1]{Corresponding author.}
  \affiliation[dsv]{organization={MIDAS, Department of Computer and Systems Sciences, Stockholm University},
    addressline={\ROMANAddressLine},
    postcode={\ROMANPostcode},
    city={Kista},
    country={Sweden}}
}

\journal{Machine Learning with Applications}

\newcommand{\ROMANAppendix}{%
  \appendix
  \numberwithin{equation}{section}%
  \numberwithin{table}{section}%
  \numberwithin{figure}{section}%
  \renewcommand{\theequation}{\Alph{section}.\arabic{equation}}%
  \renewcommand{\thetable}{\Alph{section}.\arabic{table}}%
  \renewcommand{\thefigure}{\Alph{section}.\arabic{figure}}%
}

\begin{document}
\begin{frontmatter}
\title{ROMAN: A Multiscale Routing Operator for Convolutional Time Series Models}
\ROMANAuthorBlock

\begin{abstract}
We introduce ROMAN (ROuting Multiscale representAtioN), a deterministic operator for time series that maps temporal scale and coarse temporal position into an explicit channel structure while reducing sequence length. ROMAN builds an anti-aliased multiscale pyramid, extracts fixed-length windows from each scale, and stacks them as pseudochannels. ROMAN provides a training-free mechanism to control the inductive bias of downstream models: it can reduce temporal invariance, make temporal pooling implicitly coarse-position-aware, and expose multiscale interactions through channel mixing, while often improving computational efficiency by shortening the processed time axis. We formally analyze the operator and evaluate it as preprocessing for four convolutional classifiers: MiniRocket, MultiRocket, a standard CNN-based classifier, and a fully convolutional network (FCN) classifier. Synthetic mechanism studies that isolate coarse position, long-range correlation, multiscale interaction, and full positional invariance show that ROMAN behaves as intended, and comparisons with coordinate augmentation and single-scale routing distinguish the benefits of positional routing from those of the multiscale cascade across the evaluated backbones. On long-sequence subsets of the UCR and UEA archives, ROMAN's effect on accuracy varies across datasets while its effect on efficiency is often favorable. At \(S=4\) on UCR, accuracy changes are positively correlated across the pooled backbones, consistent with the proposed role of task structure. We give practical guidance for choosing the operator's two hyperparameters and illustrate its use on very long sequences, where it is both more accurate and several times cheaper, and as a source of representation diversity for ensembles. The ROMAN operator library is available at \href{https://github.com/gon-uri/ROMAN}{github.com/gon-uri/ROMAN}.
\end{abstract}

\begin{keyword}
time series classification \sep convolutional models \sep multiscale representations \sep ROCKET \sep efficiency
\end{keyword}
\end{frontmatter}

\section{Introduction}

Convolutional methods are among the most successful approaches to time series classification (TSC) because they combine local pattern extraction, scalability, and a relatively simple optimization problem \citep{bake_off_uni_2,bake_off_multi}. This includes both learned convolutional networks and the ROCKET family of random-convolution classifiers \citep{ref_rocket}. Their strength is tied to a particular inductive bias: they are highly effective at detecting whether a local motif occurs, but less explicit about where that motif occurs. This bias is induced by the subsequent application of convolutions and pooling operations along the temporal dimension \citep{lecun_conv}. 

This limitation is especially visible when convolutions are followed by global temporal aggregation, which is a common practice in many TSC models \citep{ref_rocket,ref_inceptiontime,ref_fcn,ref_minirocket}. Global pooling is one of the main reasons convolutional models are robust and efficient, but it also suppresses temporal layout. The resulting representation is naturally shift-tolerant, which is desirable when translational invariance matches the task, but restrictive when the arrangement of events is itself discriminative. Simply removing pooling is not a satisfying alternative: flatten-based models preserve more temporal indexing, but they often sacrifice the simplicity and robustness that make pooled convolution attractive in the first place.

\begin{figure}[t]
\centering
\includegraphics[width=\textwidth]{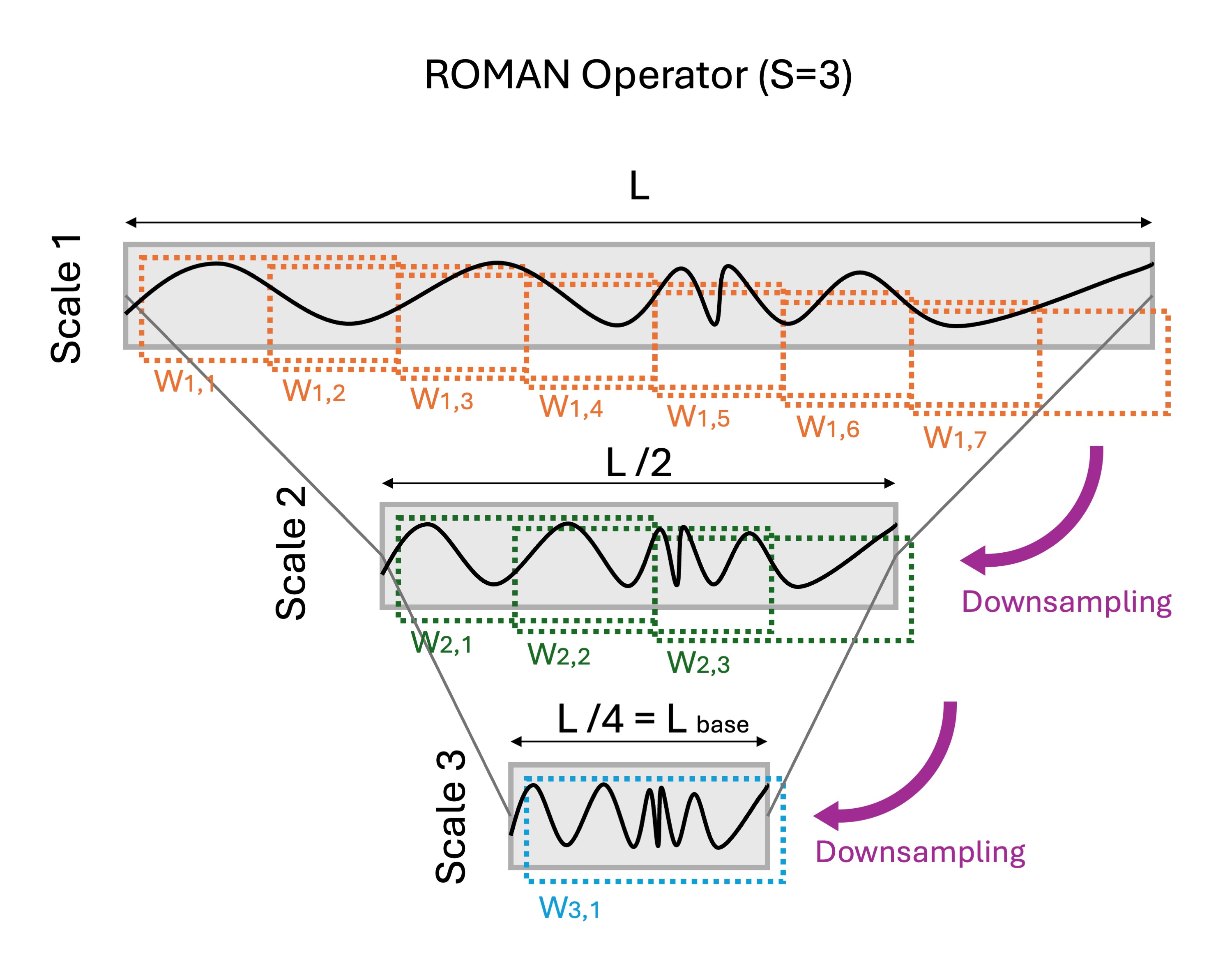}
\caption{Multiscale routing in ROMAN, illustrated for \(S=3\). Successive low-pass filtering and factor-two downsampling produce progressively coarser versions of the input. At each scale \(s\), overlapping windows of common length \(L_{\mathrm{base}}\) are extracted, with the window index \(w\) identifying coarse temporal position. Stacking these windows along the channel axis makes scale and coarse position explicit to the downstream model. In the illustrated dyadic example with \(\alpha=0.5\), the three scales contribute seven, three, and one windows, respectively.}
\label{fig:roman_overview}
\end{figure}

We address this trade-off with ROMAN, a deterministic front-end operator that reroutes temporal structure before the downstream classifier sees the input. ROMAN builds an anti-aliased multiscale pyramid, extracts fixed-length windows at each scale, and stacks them as pseudochannels. In the transformed representation, scale and coarse temporal position become explicit channel structure, while the processed temporal axis becomes shorter. Fig.~\ref{fig:roman_overview} illustrates the transformation. A downstream convolutional model can therefore remain local in time while operating on a representation that is more explicitly multiscale and more coarse-position-aware after pooling. ROMAN is not a classifier and is not intended as an architecture replacement; it is a representation operator that can be inserted before standard convolutional backbones.

The aim of this article is to study whether the multiscale routing applied by the ROMAN operator provides a useful way to modify the inductive bias of convolutional time series models. Because \(S=1\) is the original input, varying \(S\) yields a family of complementary representations that trade translational invariance, temporal awareness, multiscale coupling, and computational cost against one another. The rest of the paper develops this perspective through formal analysis and targeted empirical evaluation.

For practical use, the ROMAN library provides an installable operator with a scikit-learn-style interface and worked classification and ensemble examples: \href{https://github.com/gon-uri/ROMAN}{github.com/gon-uri/ROMAN}.

\paragraph{\textbf{Contributions}}
The paper makes an operator-level contribution rather than an architecture-level one. We introduce ROMAN as a deterministic representation map that can be combined with standard convolutional backbones and studied through the inductive biases it induces. The main contributions are:

\begin{enumerate}
    \item We formalize ROMAN as a deterministic multiscale routing operator for time series, in which the number of scales \(S\) acts as a single knob from the untouched input (\(S=1\)) to increasingly routed and fragmented representations.

    \item We analyze the transform conceptually and computationally, characterizing how ROMAN rewrites temporal scale and coarse temporal position into pseudochannels and how this changes the input size and complexity seen by MiniRocket, MultiRocket, CNNClassifier, and FCNClassifier.

    \item We construct four synthetic mechanism studies that isolate the regimes ROMAN is intended to affect---coarse temporal position, long-range correlation, multiscale interaction, and full positional invariance---and compare the operator with coordinate augmentation and single-scale routing to distinguish the benefits of positional routing from those of the multiscale cascade across the evaluated backbones.

    \item We evaluate ROMAN on long-sequence subsets of the UCR and UEA archives with pooled and position-sensitive backbones, reporting predictive performance, computational efficiency, formal paired tests, and a cross-backbone analysis supporting the hypothesis that ROMAN's benefits depend jointly on dataset structure and the information retained by the backbone.

    \item We provide practical guidance for choosing \(S\) and \(\alpha\), including an \(\alpha\)-sensitivity study on the full UCR population, and illustrate the operator's use on very long sequences and as a source of representation diversity for ensembles.
\end{enumerate}

Taken together, these results show that ROMAN's multiscale routing acts as a controllable knob on the inductive bias of convolutional models. The efficiency gain comes with the operator; the accuracy gain depends on the task, and the mechanism studies help anticipate when it is likely.

\section{Related Work and Positioning}

\paragraph{\textbf{Convolutional Backbones and Inductive Bias}}
Two families of convolutional TSC models are relevant here \citep{bake_off_uni_2,bake_off_multi,bake_off_uni}: learned architectures that fit temporal filters end-to-end, such as FCNClassifier \citep{ref_fcn}, CNNClassifier \citep{ref_conv_review}, and InceptionTime \citep{ref_inceptiontime}, and ROCKET-style models that use large collections of fixed or random kernels followed by a simple classifier, such as ROCKET, MiniRocket, MultiRocket, and Detach-Rocket \citep{ref_rocket,ref_minirocket,ref_multirocket,ref_detach}. For the purposes of this paper, the key distinction between these models is not convolution alone, but the temporal information retained after convolution. ROCKET and MiniRocket are strongly pooling-based and therefore emphasize motif occurrence while largely attenuating absolute temporal location. Later iterations enrich this representation in different ways: MultiRocket introduces additional pooling operators, whereas HDC-MiniROCKET incorporates temporal encoding through hyperdimensional computing \citep{ref_multirocket,ref_hdc_minirocket}. Both build on the random-convolution approach. FCNClassifier follows a learned variant of the same principle through global average pooling, whereas CNNClassifier ends with a flattened temporal representation and therefore preserves substantially more positional indexing. This distinction is central for interpreting ROMAN: when class information depends on coarse temporal location, pooled and position-sensitive backbones should not be expected to respond in the same way.

A second relevant distinction concerns how convolutional models access long-range temporal structure. In ROCKET-style methods, coverage over multiple temporal scales is obtained by using kernels with a wide range of dilation values, from highly local to effectively global ones \citep{ref_rocket,ref_minirocket,ref_multirocket}. This provides broad receptive-field coverage, but the same kernel does not naturally bind multiple distant local events while preserving detailed local structure. In learned convolutional networks, long-range dependencies are typically built through depth: successive convolutional layers enlarge the effective receptive field and progressively relate local evidence over longer temporal spans \citep{lecun_conv,ref_inceptiontime}. For long sequences, however, capturing relations between distant local events may require substantial network depth before they fall within the effective receptive field, and such depth can make optimization more challenging.

\paragraph{\textbf{Changing the Model versus Changing the Representation}}
Methods that give convolutional models better access to temporal scale or position modify the architecture, the input representation, or both. Causal dilated convolutions \citep{ref_Bai} and parallel multi-branch kernels \citep{ref_inceptiontime,ref_Tang,ref_prism} widen scale coverage, anti-aliased downsampling reduces sensitivity to small shifts \citep{shift_invariance}, and TimesNet folds the series into a two-dimensional representation according to estimated dominant periods \citep{ref_timesnet}. These approaches differ in whether they learn the representation and whether they reduce temporal resolution. ROMAN instead uses a fixed input transformation to expose scale and coarse position jointly through channel structure.

Multi-scale Convolutional Neural Networks (MCNN) also construct downsampled and smoothed versions of the input, which are processed through learned convolutional branches and subsequently combined \citep{ref_mcnn}. In MCNN, multiscale processing is integrated into the CNN architecture through these branches and their learned fusion. ROMAN differs by routing windows from multiple scales onto a common temporal axis before feature extraction, allowing the resulting representation to be supplied to different downstream backbones.

Among deterministic front-ends, wavelet scattering produces nonlinear multiscale features whose averaging reduces sensitivity to local translations \citep{ref_scattering,ref_scattering_audio}. ROMAN instead routes signal windows into channels indexed by scale and coarse temporal location, retaining waveform samples at the finest level. This exposes local waveform structure and provides a mechanism for preserving coarse positional distinctions through downstream temporal pooling---a routing mechanism that standard scattering does not supply. Patch-based tokenization and space-to-depth rearrangements \citep{ref_patchtst,ref_spd} stack windows of a single temporal scale; our synthetic ablation shows that adding coarser scales can substantially improve access to cross-scale structure for MiniRocket and MultiRocket (Sect.~\ref{sec:synthetic}). CoordConv appends coordinate channels to the input \citep{ref_Liu}, but the ability to exploit these coordinates depends on the downstream model: in our position-awareness experiment, they are effective for FCNClassifier but do not reproduce the gains from routing for the tested ROCKET pipelines; HDC-MiniROCKET encodes time into the feature space of one specific model family \citep{ref_hdc_minirocket}. Within this landscape, we introduce a fixed routing construction that maps windows from multiple levels of a low-pass pyramid onto a common, shorter time axis, encoding their scale and coarse temporal location through channel identity. This provides a training-free front-end that can be evaluated across different downstream backbones without changing their architectures.

\paragraph{\textbf{Positioning of ROMAN}}
ROMAN is a deterministic front-end operator rather than a new classifier architecture: it does not replace downstream models or rely on learned task-specific filters. This separation is scientifically useful because it enables representation-level analysis while keeping the downstream classifier fixed: at the operator level, the questions concern the inductive biases introduced by multiscale routing as \(S\) grows from the identity case \(S=1\); at the model level, how standard backbones behave on the transformed input. ROMAN is therefore best understood not as a competitor to the strong classifiers that lead current benchmark evaluations, such as Hydra and HIVE-COTE~2.0 \citep{ref_hydra,ref_hive,bake_off_uni_2}, but as an orthogonal, combinable way to expose temporal structure that pooled convolution suppresses.

\section{ROMAN: Definition and Properties}

\subsection{Definition}

Let \(\mathbf{X}\in\mathbb{R}^{C\times L}\) be a multivariate time series with \(C\) channels and temporal length \(L\). For fixed scale depth \(S\geq 1\) and overlap parameter \(\alpha\in[0,1)\), ROMAN constructs a dyadic anti-aliased pyramid, where \(s=1\) denotes the finest level and \(s=S\) the coarsest one. Define
\[
\mathbf{X}^{(1)}=\mathbf{X},
\qquad
\mathbf{X}^{(s)}=\mathcal{D}_2\!\left(h * \mathbf{X}^{(s-1)}\right),\quad s=2,\ldots,S,
\]
where \(h=\tfrac14[1,2,1]\), \( * \) denotes channelwise one-dimensional convolution with fixed boundary handling, and \(\mathcal{D}_2\) denotes decimation by a factor of two. Let \(L_s\) be the length of \(\mathbf{X}^{(s)}\), and set
\[
L_{\mathrm{base}} := L_S.
\]
Thus all routed windows share the common temporal length \(L_{\mathrm{base}}\). Following the anti-aliased downsampling principle used in BlurPool \citep{shift_invariance}, we apply a low-pass filter before temporal subsampling to attenuate high-frequency components that would otherwise alias. This is intended to improve the stability of the coarser representations to small temporal shifts. We selected the binomial filter \(h=\tfrac14[1,2,1]\) to satisfy the discrete scale-space axioms, guaranteeing no new extrema are created during smoothing \citep{filter}.

At each scale \(s\), ROMAN extracts windows of length \(L_{\mathrm{base}}\). The number of windows is
\[
W_s
=
1+\left\lceil
\frac{L_s-L_{\mathrm{base}}}{(1-\alpha)L_{\mathrm{base}}}
\right\rceil,
\]
and the corresponding start indices are
\[
a_{s,w}=
\begin{cases}
1, & W_s=1,\\[4pt]
1+\left\lfloor\dfrac{(w-1)(L_s-L_{\mathrm{base}})}{W_s-1}\right\rfloor, & W_s>1,
\end{cases}
\qquad
w=1,\ldots,W_s.
\]
The extracted window at scale \(s\) and position \(w\) is
\[
\mathbf{U}^{(s,w)}
=
\mathbf{X}^{(s)}_{:,\,a_{s,w}:a_{s,w}+L_{\mathrm{base}}-1}
\in\mathbb{R}^{C\times L_{\mathrm{base}}}.
\]
This placement spreads the windows uniformly over the admissible start range; the realized overlap may therefore differ from the target value \(\alpha\) only through integer rounding.

\begin{figure}[!t]
\centering
\includegraphics[width=\textwidth]{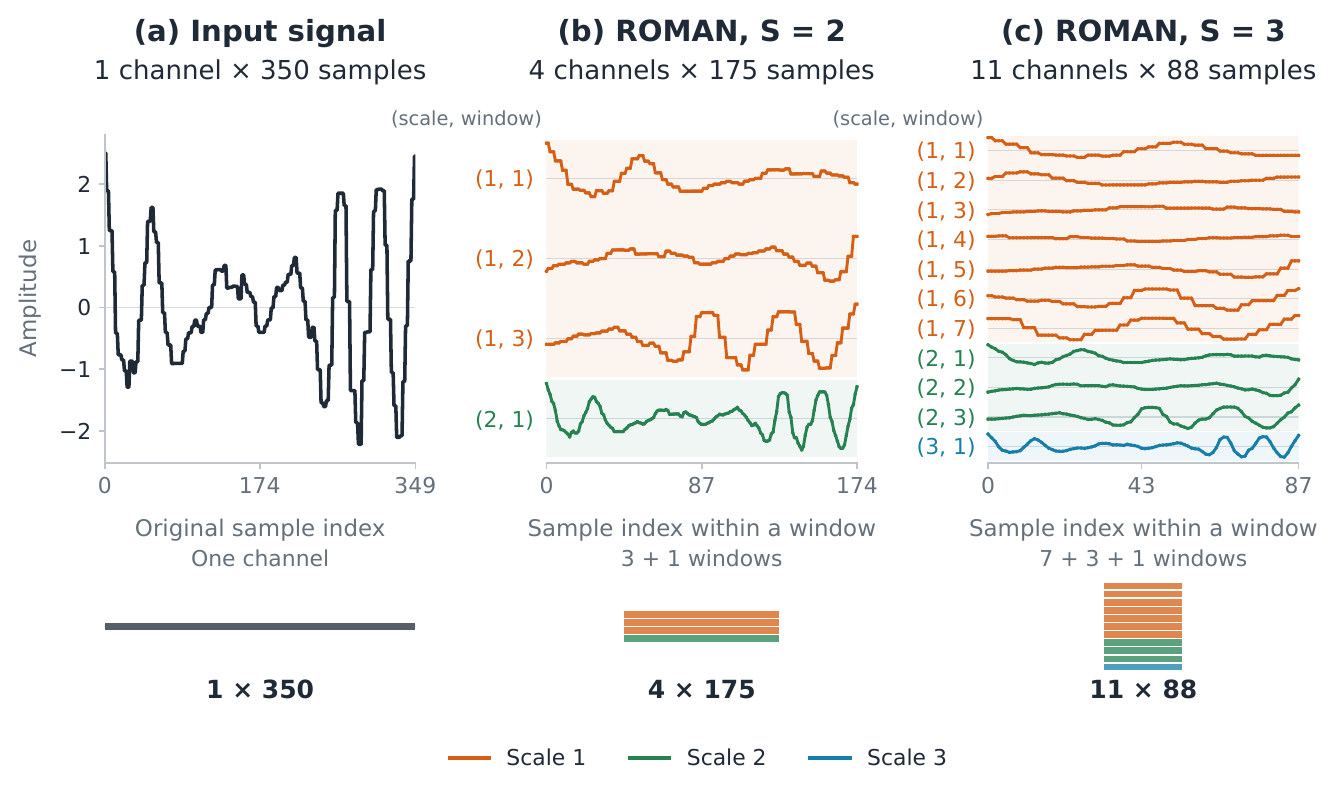}
\caption{ROMAN applied to a real time series from FaceFour with \(\alpha=0.5\). Increasing the routing depth reorganizes the same input into more channels representing different temporal regions and scales, while shortening their common time axis. Fine-scale channels retain local waveform detail, whereas coarser channels cover longer spans of the original signal. The lower diagrams illustrate this change from a long single-channel input to shorter multichannel representations.}
\label{fig:roman_real_example}
\end{figure}

The ROMAN output is obtained by concatenating all routed windows along the channel axis:
\[
\mathbf{Z}
=
\big[
\mathbf{U}^{(1,1)};
\ldots;
\mathbf{U}^{(1,W_1)};
\mathbf{U}^{(2,1)};
\ldots;
\mathbf{U}^{(S,W_S)}
\big]
\in \mathbb{R}^{C'\times L_{\mathrm{base}}},
\qquad
C' = C\sum_{s=1}^{S} W_s,
\]
where the semicolon denotes concatenation along the channel axis. We denote this map by
\[
\mathrm{ROMAN}_{S,\alpha}:\mathbb{R}^{C\times L}\to\mathbb{R}^{C'\times L_{\mathrm{base}}}.
\]
Each pseudochannel therefore corresponds to a triple consisting of an original channel, a scale, and a coarse temporal window. Fig.~\ref{fig:roman_real_example} illustrates the resulting representation on a real input for \(S=2\) and \(S=3\).

\paragraph{Roles of \(S\) and \(\alpha\)}
The two hyperparameters play different roles. The depth \(S\) determines the common base length \(L_{\mathrm{base}}\) through the pyramid depth, whereas \(\alpha\) controls how densely each scale is tiled by overlapping windows. In particular, increasing \(S\) shortens the processed temporal axis, while increasing \(\alpha\) increases representational redundancy through larger \(W_s\).

When \(S=1\), one has \(L_{\mathrm{base}}=L\), \(W_1=1\), and \(C'=C\), so
\[
\mathrm{ROMAN}_{1,\alpha}(\mathbf{X})=\mathbf{X}.
\]
That is, \(S=1\) is the identity and provides the no-ROMAN baseline used throughout the experiments, and \(S\) can be read as a knob from the untouched input to increasingly routed and fragmented representations. Exact boundary-handling details, odd-length conventions, and an equivalent parameterization in terms of a minimum admissible base length are given in \ref{app:operator}.

\subsection{Representation and Inductive Bias}
ROMAN changes the representation seen by the downstream model, not the model itself. After the transform, convolution operates on a pseudo-multivariate series in which each channel is indexed by original channel identity, pyramid scale, and coarse temporal window. Temporal scale and coarse temporal position are therefore made explicit before any learned or random convolution is applied. This has two immediate consequences: channel mixing can combine information across scales directly, and subsequent temporal pooling becomes less translation-invariant because activations from different coarse temporal regions are no longer pooled within the same channel.

ROMAN is not equivalent to simply slicing the finest-scale signal into shorter windows. The coarser pyramid levels retain low-frequency and long-range structure at reduced temporal resolution, so shrinking \(L_{\mathrm{base}}\) does not merely discard slow phenomena. In this sense, ROMAN defines a family of representations indexed by \(S\): the original input at \(S=1\), and, as \(S\) grows, a shorter processed temporal axis and a channel axis increasingly enriched with coarse location and scale information. The purpose of the paper is therefore not to claim that larger \(S\) is uniformly better, but to study when this change in inductive bias is useful.

\paragraph{Representation size and cost}
The transformed representation contains
\[
|\mathbf{Z}| = C' L_{\mathrm{base}} = C L_{\mathrm{base}}\sum_{s=1}^{S} W_s
\]
scalars. Constructing the pyramid costs \(\Theta\!\big(C\sum_{s=1}^{S} L_s\big)\), while extracting and stacking the routed windows costs \(\Theta(|\mathbf{Z}|)\). Since dyadic downsampling gives \(\sum_{s=1}^{S} L_s = \Theta(L)\), the cost of applying ROMAN to a single series is
\[
T_{\mathrm{ROMAN}} = \Theta(CL + |\mathbf{Z}|),
\]
and the dataset-level cost is linear in the number of series.

The practical effect on downstream models is the same for all four backbones: ROMAN decreases the temporal length from \(L\) to \(L_{\mathrm{base}}\) while increasing the number of channels from \(C\) to \(C'\); how this trades off against each backbone's own cost structure is derived in \ref{app:operator}.

\section{Synthetic Mechanism Studies}
\label{sec:synthetic}

Each experiment compares a standard backbone with the same backbone preceded by ROMAN; unless otherwise stated, we fix \(\alpha=0.5\) and evaluate \(S\in\{1,2,3,4\}\), where \(S=1\) is the baseline and larger values induce progressively stronger multiscale routing. The downstream backbones are MiniRocket \citep{ref_minirocket}, MultiRocket \citep{ref_multirocket}, CNNClassifier \citep{ref_conv_review}, and FCNClassifier \citep{ref_fcn} (abbreviated CNN and FCN in tables), chosen to span pooled random-convolution models, a learned pooled-convolution model, and a learned position-sensitive convolutional model; all four are used through their \texttt{sktime} implementations \citep{sktime}, and implementation details and fixed hyperparameters are given in \ref{app:environment}. The synthetic experiments test the mechanisms claimed for ROMAN under controlled conditions. They isolate four representational regimes in which coarse temporal position, long-range relations, multiscale binding, or full translational invariance are decisive. To attribute the observed effects to specific ingredients of the operator, each task is evaluated under four input representations: the raw series (Base); the raw series with a linear coordinate channel appended, i.e., a CoordConv-style input \citep{ref_Liu} that injects position without restructuring the signal (Coord); ROMAN without its multiscale cascade, i.e., only the \(W_1\) finest-scale windows of the operator stacked as channels, which routes and shortens exactly as ROMAN does but contains no coarser scale (ROMAN without multiscale); and the full operator at \(S=4\) (ROMAN). The former is a slice of the latter's output, so the two share window geometry by construction. We report representative examples and mean test accuracy over ten realizations.

\begin{figure}[!t]
\centering
\footnotesize
\setlength{\tabcolsep}{2.5pt}
\renewcommand{\arraystretch}{1.0}

\begin{minipage}[c]{0.35\columnwidth}
\centering
\includegraphics[width=\linewidth]{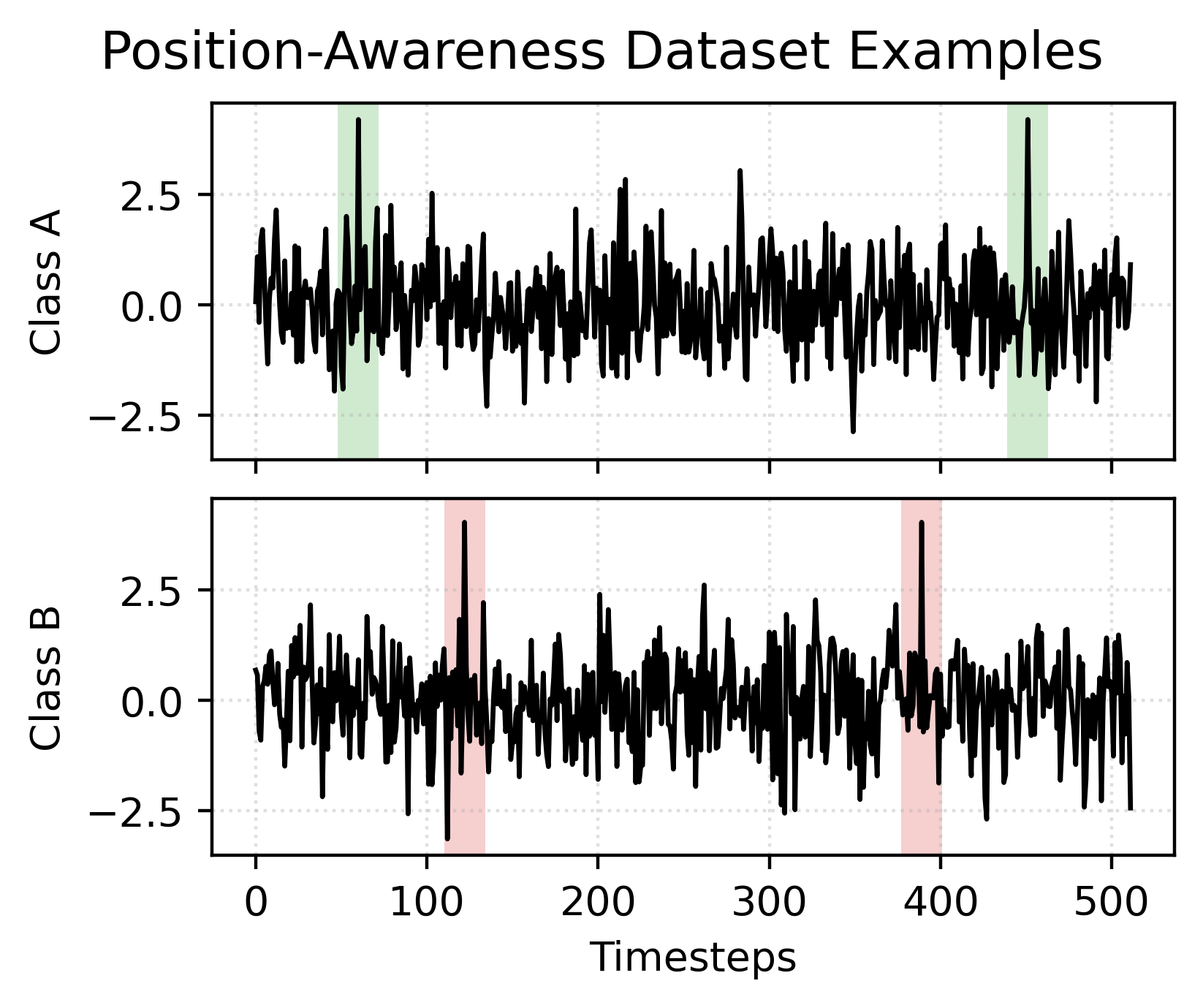}
\end{minipage}\hfill
\begin{minipage}[c]{0.63\columnwidth}
\centering
\textbf{(a) Coarse Position Awareness}\\[0.3em]
\begin{tabular}{lcccc}
\toprule
\multicolumn{1}{l}{Model} & Base & Coord & ROMAN & ROMAN \\
 & & \shortstack{\citeauthor{ref_Liu}\\(\citeyear{ref_Liu})} & (no multiscale) & (\(S{=}4\)) \\
\midrule
MiniRocket  & .71 & .72 & \underline{.90} & \textbf{.91} \\
MultiRocket & .75 & .73 & \textbf{.93} & \underline{.92} \\
CNN         & \textbf{.97} & \underline{.95} & .71 & .71 \\
FCN         & \nc{.51} & \textbf{1.00} & \underline{.84} & .79 \\
\bottomrule
\end{tabular}
\end{minipage}

\vspace{0.4em}

\begin{minipage}[c]{0.35\columnwidth}
\centering
\includegraphics[width=\linewidth]{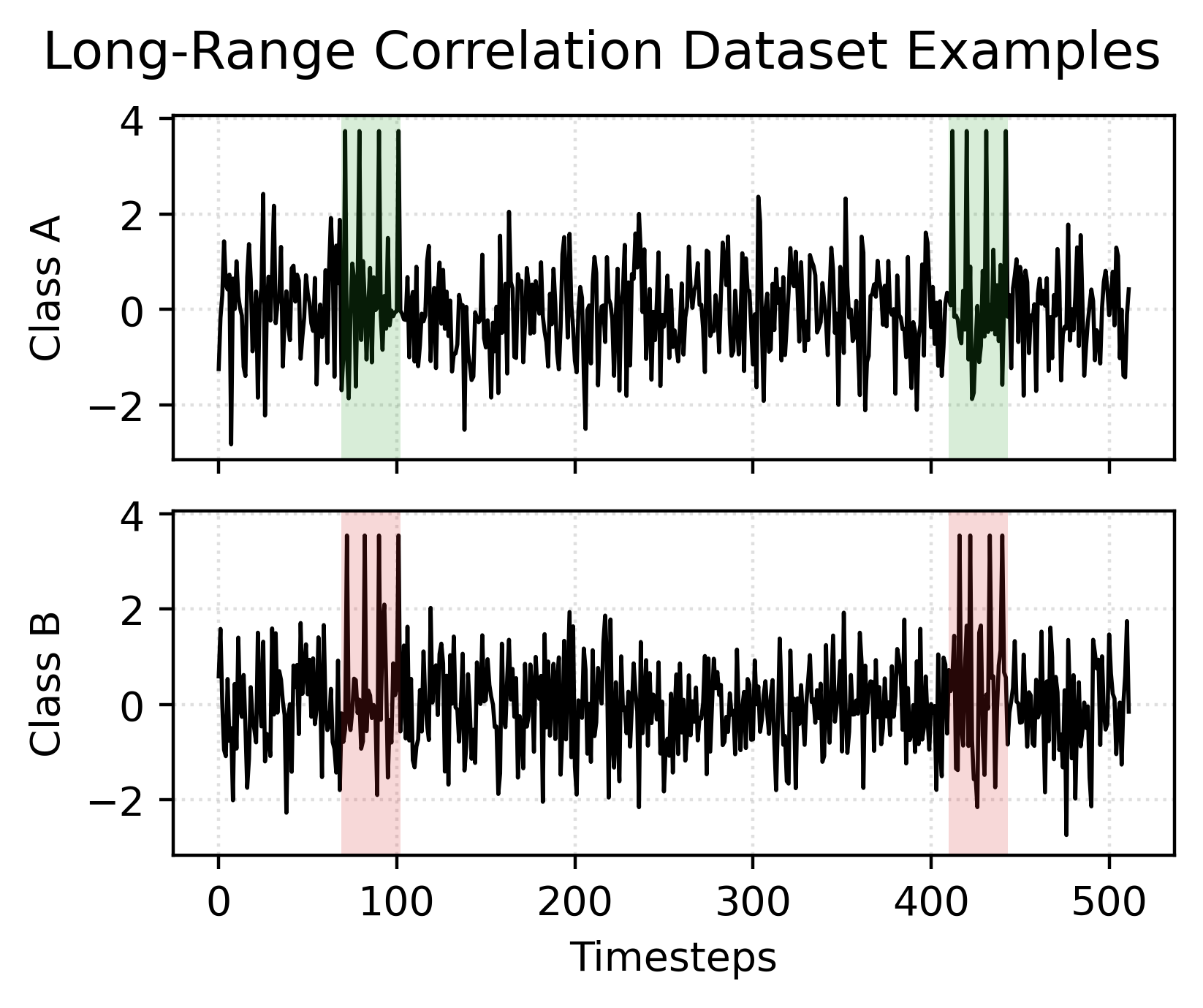}
\end{minipage}\hfill
\begin{minipage}[c]{0.63\columnwidth}
\centering
\textbf{(b) Long-Range Correlation}\\[0.3em]
\begin{tabular}{lcccc}
\toprule
\multicolumn{1}{l}{Model} & Base & Coord & ROMAN & ROMAN \\
 & & \shortstack{\citeauthor{ref_Liu}\\(\citeyear{ref_Liu})} & (no multiscale) & (\(S{=}4\)) \\
\midrule
MiniRocket  & \nc{.51} & \nc{.52} & \textbf{.71} & \underline{.68} \\
MultiRocket & .56 & .54 & \textbf{.72} & \underline{.68} \\
CNN         & \nc{.51} & \nc{.51} & \underline{.90} & \textbf{.91} \\
FCN         & \nc{\underline{.50}} & \nc{.49} & \textbf{1.00} & \textbf{1.00} \\
\bottomrule
\end{tabular}
\end{minipage}

\caption{Synthetic mechanism studies for (a) coarse position awareness and (b) long-range correlation. Left: representative examples. Right: mean test accuracy over ten realizations under the four input representations: raw series (Base), raw series with a coordinate channel (Coord), ROMAN without its multiscale cascade, i.e., its scale-1 windows only, and full ROMAN at \(S=4\); standard deviations are reported in \ref{app:ablation}. Bold and underlining mark the highest and second-highest distinct mean accuracies, respectively, at the displayed precision in each row; ties receive the same marking. These rankings are descriptive and do not indicate statistical equivalence. Gray marks accuracies that do not exceed the independent random-guessing reference at the 5\% level (one-sided binomial test with \(p_0=0.50\), pooling 2,500 test predictions across ten realizations; uncorrected).}
\label{fig:synthetic_summary_1}
\end{figure}

\paragraph{\textbf{a) Coarse Position Awareness}}
The first synthetic family isolates the simplest mechanism targeted by ROMAN: coarse temporal position can be discriminative even when local motif content is identical. Each series contains two identical spikes placed symmetrically with respect to the sequence boundaries, and the class label depends only on their coarse distance from the borders. Because the spikes are identical and mirror-symmetric, any feature that is linear in position is constant across classes, so the task isolates coarse positional information and cannot be solved from motif identity or from a simple positional average.

CNNClassifier is expected to perform strongly without ROMAN because its flattened head preserves temporal indexing. The more informative comparison is between pooled models and their ROMAN-augmented counterparts: if ROMAN improves MiniRocket, MultiRocket, or FCNClassifier in this setting, it indicates that routing coarse temporal position into pseudochannels restores information that temporal pooling would otherwise suppress. The results in Fig.~\ref{fig:synthetic_summary_1}(a) are consistent with this mechanism.

\paragraph{\textbf{b) Long-Range Correlation}}
The second synthetic family tests whether classification depends on a relation between two distant localized events rather than on the presence of either event alone. Two sparse burst windows are placed at fixed, widely separated positions, and the class label depends on whether the two bursts contain the same local pattern or different local patterns; since the burst geometry is identical across classes, neither position nor the presence of either burst alone carries information.

This task highlights a known tension in convolutional models: relating the two bursts requires a receptive field large enough to span both regions, but increasing dilation also makes it harder to preserve fine local structure. ROMAN is intended to relax this trade-off by routing distant regions into different pseudochannels, so that shorter-range convolutions can combine them after transformation. The results in Fig.~\ref{fig:synthetic_summary_1}(b) support this.

\begin{figure}[!t]
\centering
\footnotesize
\setlength{\tabcolsep}{2.5pt}
\renewcommand{\arraystretch}{1.0}

\begin{minipage}[c]{0.35\columnwidth}
\centering
\includegraphics[width=\linewidth]{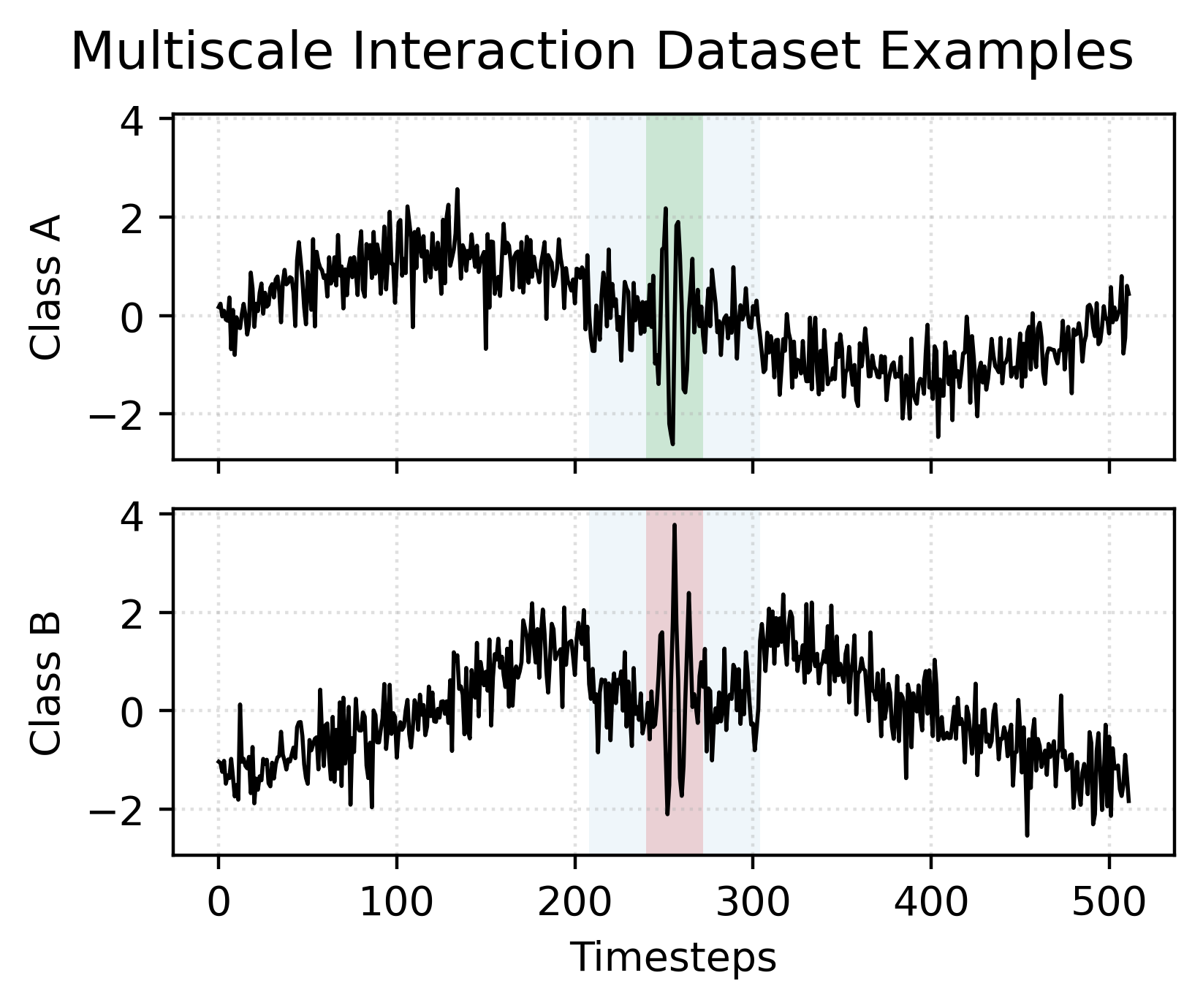}
\end{minipage}\hfill
\begin{minipage}[c]{0.63\columnwidth}
\centering
\textbf{(c) Multiscale Interaction}\\[0.3em]
\begin{tabular}{lcccc}
\toprule
\multicolumn{1}{l}{Model} & Base & Coord & ROMAN & ROMAN \\
 & & \shortstack{\citeauthor{ref_Liu}\\(\citeyear{ref_Liu})} & (no multiscale) & (\(S{=}4\)) \\
\midrule
MiniRocket  & .55 & .54 & \underline{.58} & \textbf{.73} \\
MultiRocket & .54 & .54 & \underline{.58} & \textbf{.74} \\
CNN         & \nc{.50} & \nc{.50} & \underline{.99} & \textbf{1.00} \\
FCN         & \nc{\underline{.50}} & \nc{\underline{.50}} & \textbf{1.00} & \textbf{1.00} \\
\bottomrule
\end{tabular}
\end{minipage}

\vspace{0.4em}

\begin{minipage}[c]{0.35\columnwidth}
\centering
\includegraphics[width=\linewidth]{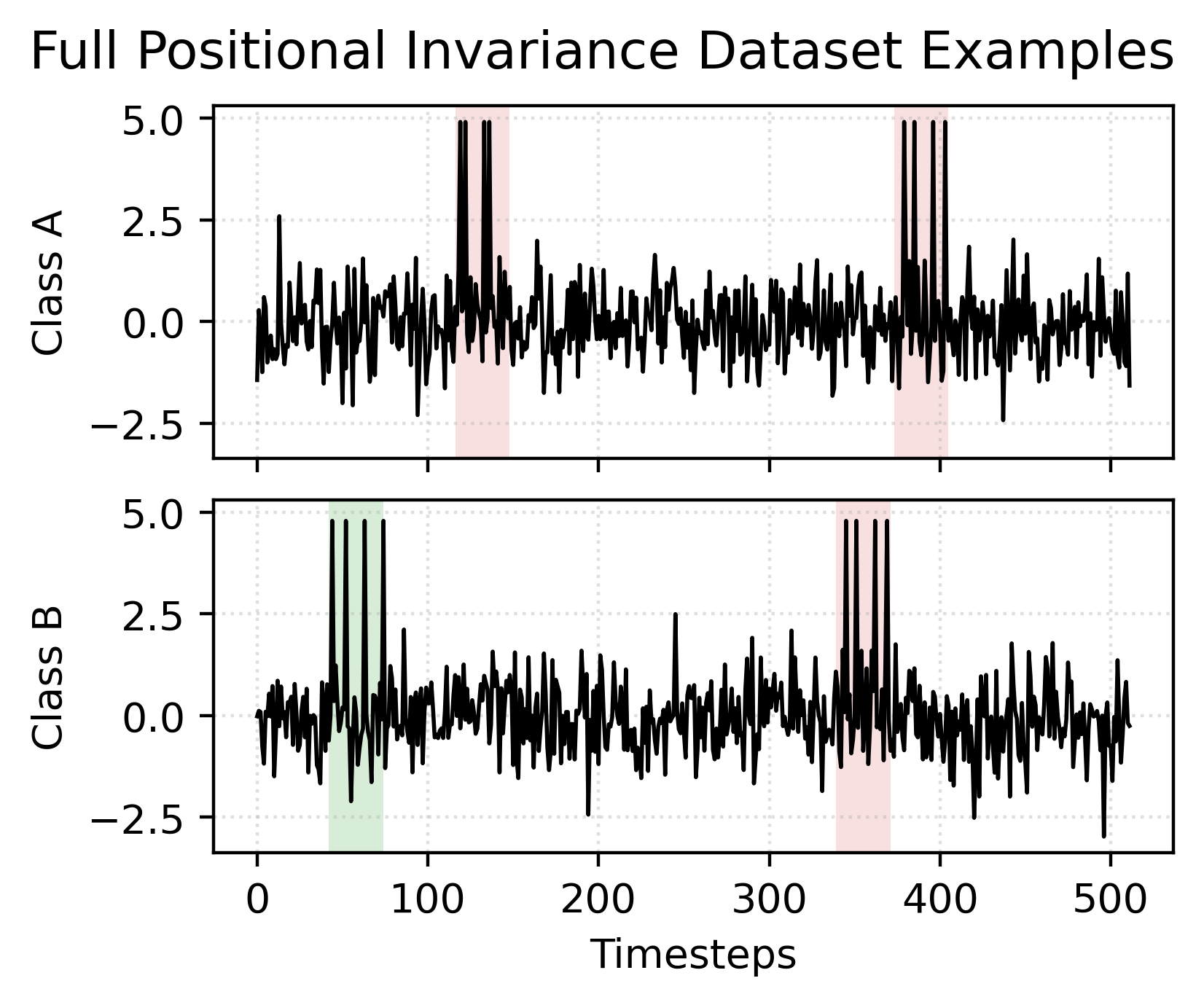}
\end{minipage}\hfill
\begin{minipage}[c]{0.63\columnwidth}
\centering
\textbf{(d) Full Positional Invariance}\\[0.3em]
\begin{tabular}{lcccc}
\toprule
\multicolumn{1}{l}{Model} & Base & Coord & ROMAN & ROMAN \\
 & & \shortstack{\citeauthor{ref_Liu}\\(\citeyear{ref_Liu})} & (no multiscale) & (\(S{=}4\)) \\
\midrule
MiniRocket  & \textbf{.97} & \underline{.96} & .68 & .65 \\
MultiRocket & \textbf{.84} & \underline{.78} & .62 & .63 \\
CNN         & \nc{\underline{.50}} & \nc{\textbf{.51}} & \nc{.49} & \nc{.49} \\
FCN         & \textbf{1.00} & \textbf{1.00} & \nc{.51} & \underline{.52} \\
\bottomrule
\end{tabular}
\end{minipage}

\caption{Synthetic mechanism studies for (c) multiscale interaction and (d) full positional invariance. Left: representative examples. Right: mean test accuracy over ten realizations under the same four input representations and marking conventions as Fig.~\ref{fig:synthetic_summary_1}.}
\label{fig:synthetic_summary_2}
\end{figure}

\paragraph{\textbf{c) Multiscale Interaction}}
The third synthetic family targets dependence between fine-scale and coarse-scale evidence at the same nominal temporal location. The construction combines a low-frequency cosine spanning the full series, a localized high-frequency burst, and a masked central region that hides the coarse phase locally, so that the coarse phase can be read only from the outer context and never from the burst's own support. The class label depends on whether the fine burst phase matches the coarse-scale phase. A local detector can resolve the burst phase but not the surrounding coarse context, whereas a wide-scale detector can infer the coarse phase from the outer signal but does not naturally bind it to the local burst. This is a direct synthetic test of ROMAN as an explicit multiscale representation. In this setting, ROMAN makes fine- and coarse-scale evidence jointly accessible through channel mixing on a shorter temporal axis. Fig.~\ref{fig:synthetic_summary_2}(c) shows that the full pyramid substantially improves MiniRocket and MultiRocket over routing alone, while CNNClassifier and FCNClassifier already achieve near-perfect accuracy with the single-scale routing condition.

\paragraph{\textbf{d) Full Positional Invariance}}
The fourth synthetic family serves as a negative control in which full translational invariance is genuinely desirable. In this case, the class label depends solely on the presence of a target motif, which can occur at any temporal location alongside similar distractor motifs; positive and negative series contain the same number of motif instances, so only motif identity, not count or position, separates the classes. Therefore, the task aligns naturally with pooling-based architectures that emphasize motif occurrence and suppress absolute position. This experiment tests the limits of ROMAN's usefulness. If the discriminative structure is fully shift-invariant, routing coarse temporal position into pseudochannels is counterproductive because it weakens an invariance that already matches the task. Strong baseline performance from MiniRocket, MultiRocket, and FCNClassifier is therefore expected. The pattern in Fig.~\ref{fig:synthetic_summary_2}(d) is consistent with this interpretation.

\paragraph{\textbf{e) Which Ingredient Does the Work}}
Across the four tasks, the ablations show that the contribution of each of ROMAN's ingredients depends on both the task and the downstream backbone. On the position and long-range tasks, the finest-scale windows capture most of the benefit where routing improves performance; adding coarser scales provides little further advantage and sometimes lowers accuracy. On the multiscale-interaction task, however, the pyramid raises MiniRocket and MultiRocket from approximately 58\% accuracy with routing alone to 73--74\%. CNNClassifier and FCNClassifier already reach 99--100\% without the pyramid, showing that learned convolutional backbones can recover the relevant dependence from the routed single-scale representation. Explicit coordinates provide a different intervention: they solve the position-awareness task for FCNClassifier but do not reproduce the routing gains for the ROCKET models. Finally, the negative control exposes the cost of weakening a useful invariance. Routing strongly harms the three backbones that perform well on the raw input, whereas coordinate augmentation has smaller effects, including a six-percentage-point loss for MultiRocket. Together, these results support ROMAN as a way to expose temporal structure whose usefulness depends on what the backbone can already extract.

\section{Archive Benchmarks on UCR and UEA}

We evaluate ROMAN on the UCR and UEA archives \citep{ref_ucr,ref_uea}, restricting attention to datasets with sequence length \(L \geq 256\) so that after applying ROMAN with \(S=4\) the reduced representation still has sufficient temporal extent for the downstream kernels to operate (\(L_{\mathrm{base}} \geq 32\)). For the early-stopping convolutional models, a small number of long UCR datasets are excluded when severe class imbalance prevents a representative validation split; the exact exclusion list is reported in \ref{app:archive}.

\subsection{UCR results}
\label{sec:ucr}
\begin{figure}[t]
\centering
\includegraphics[width=\textwidth]{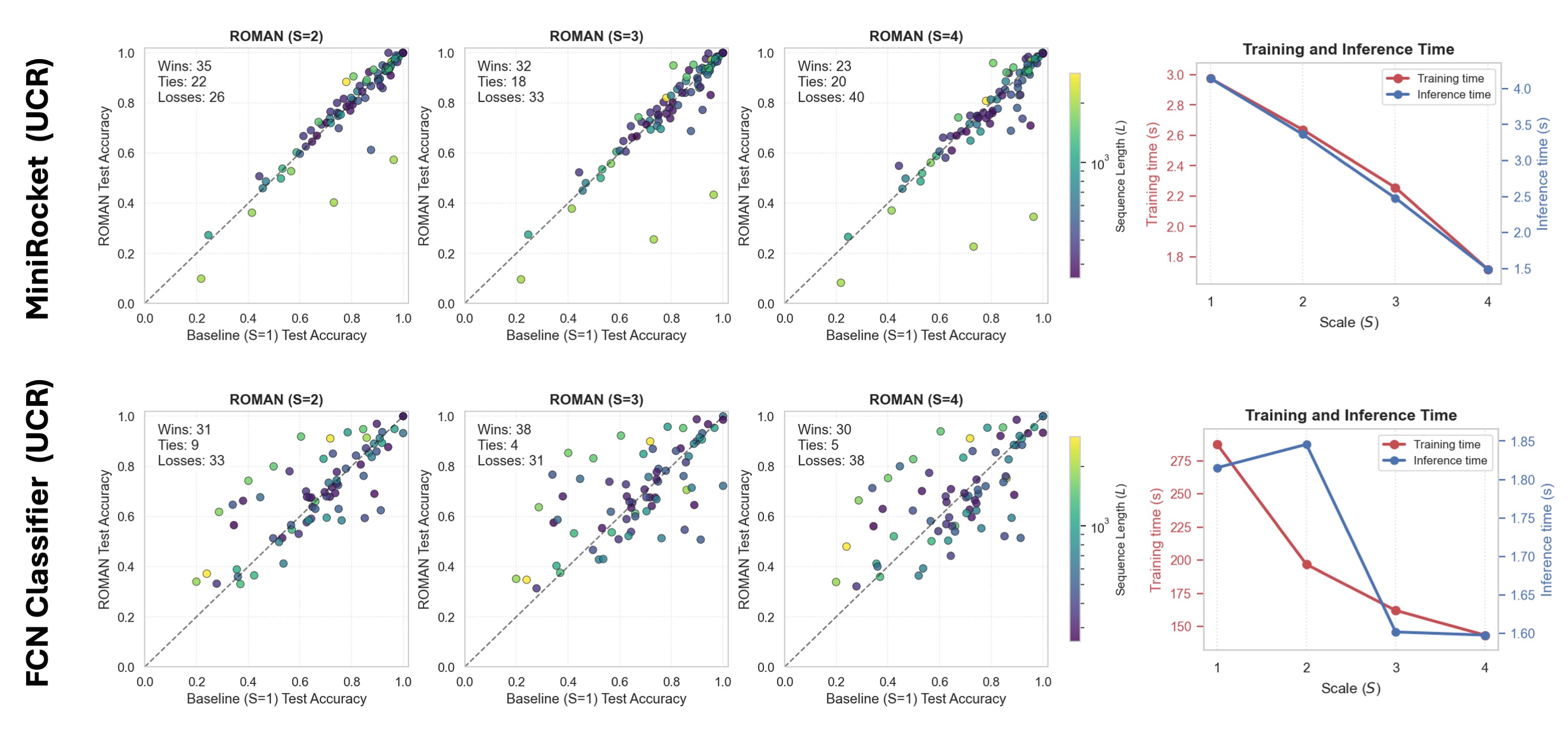}
\caption{Baseline (\(S = 1\)) vs. ROMAN-transformed data (\(S \in \{2,3,4\}\)) results for the MiniRocket and FCN Classifier models. We present the per-dataset UCR accuracies and the scaling of computational times with respect to \(S\).}
\label{fig:UCR_results}
\end{figure}

Table \ref{tab:ucr_archive_summary} summarizes the effect of ROMAN on the long-sequence UCR benchmark. For each backbone and each nontrivial configuration \(S \in \{2,3,4\}\), we report wins, ties, losses, accuracy differences, and relative training and inference times against the corresponding \(S=1\) baseline. We also report the absolute accuracy-difference to summarize the typical magnitude of the change across datasets. Fig.~\ref{fig:UCR_results} complements this summary with per-dataset results for MiniRocket and FCNClassifier together with the computational time scaling as a function of \(S \).

The archive-level view supports a balanced reading: ROMAN is not uniformly beneficial, and no single \(S\) dominates. Formal paired tests confirm this picture (Wilcoxon signed-rank with Holm correction, \ref{app:stats}): most backbone--configuration comparisons are statistically indistinguishable from the \(S=1\) baseline, CNNClassifier is significantly better with ROMAN at every \(S\) (about \(+1\) percentage point, pp), and MultiRocket is significantly worse. One possible explanation is that MultiRocket's existing temporal summaries \citep{ref_multirocket} leave less room for routing to help, while changes to invariance and input geometry can still reduce accuracy.

Aggregate statistics, however, average over exactly the heterogeneity that the operator is designed to expose, and the more informative question is \emph{which} datasets benefit. Two facts stand out. First, benefits are far from rare: a ROMAN configuration beats the \(S=1\) baseline by more than 0.5 pp on 47\% of the datasets for MiniRocket, 41\% for MultiRocket, 77\% for CNNClassifier and 59\% for FCNClassifier, so the representation is complementary to the baseline rather than a replacement for it. Second, at parity of accuracy the transform is often simply cheaper, because it rewrites a long temporal axis into a shorter one before the classifier is applied: at \(S=4\), MiniRocket and MultiRocket reach median inference-time ratios of 0.40 and 0.30, and on roughly half of the datasets (51\% and 42\%) this speed-up comes at no accuracy cost. ROMAN is therefore not free---it expands the channel axis---but it often trades that expansion for a more valuable reduction in processed sequence length (\ref{app:operator}).

The mechanism studies suggest that ROMAN's effect may depend on an interaction between task structure and the information retained by the backbone. Table~\ref{tab:cross_backbone} examines whether different backbones respond similarly across the 73 shared UCR datasets. At \(S=4\), accuracy changes are positively correlated among the three pooled backbones (\(\rho=0.34\)--\(0.61\)), whereas the correlations involving CNNClassifier are small (\(\rho=-0.13\)--\(0.12\)). This pattern is consistent with routing exposing information that pooled representations make less accessible. CNNClassifier's different response may reflect its preservation of temporal indexing and the changes ROMAN introduces to its input geometry. The four synthetic tasks (Sect.~\ref{sec:synthetic}) illustrate its distinct response: ROMAN helps CNNClassifier on long-range correlation and multiscale interaction, where it performs similarly with and without the multiscale cascade; it solves the position task without ROMAN and loses accuracy with it; and its accuracy remains near chance on the invariance task, with or without ROMAN.

\begin{table}[t]
\centering
\caption{Archive-level summary of ROMAN on the long-sequence UCR benchmark. For each backbone and each nontrivial ROMAN configuration \(S \in \{2,3,4\}\), the table reports wins, ties, and losses against the corresponding \(S=1\) baseline. The accuracy-difference columns are reported across datasets in percentage points as median \([Q_1,Q_3]\). The final two columns report median per-dataset training and inference time ratios relative to baseline.}
\label{tab:ucr_archive_summary}
\begingroup
\setlength{\tabcolsep}{3.5pt}
\resizebox{\textwidth}{!}{%
\begin{tabular}{llccccccc}
\toprule
Model & Config. & Wins & Ties & Losses & Acc. diff. & Absolute acc. diff. & Train ratio & Infer. ratio \\
\midrule
MiniRocket & S=2 & 35 & 22 & 26 & \gooddelta{+0.11 [-0.89, +2.09]} & 1.57 [0.43, 4.01] & \gooddelta{$\times$0.88} & \gooddelta{$\times$0.84} \\
MiniRocket & S=3 & 32 & 18 & 33 & \gooddelta{+0.00 [-2.12, +2.17]} & 2.29 [0.62, 4.01] & \gooddelta{$\times$0.73} & \gooddelta{$\times$0.63} \\
MiniRocket & S=4 & 23 & 20 & 40 & \baddelta{-0.44 [-3.86, +0.70]} & 2.14 [0.61, 4.42] & \gooddelta{$\times$0.59} & \gooddelta{$\times$0.40} \\
\midrule
MultiRocket & S=2 & 23 & 23 & 37 & \baddelta{-0.33 [-1.48, +0.62]} & 1.00 [0.45, 2.32] & \gooddelta{$\times$0.77} & \gooddelta{$\times$0.71} \\
MultiRocket & S=3 & 22 & 21 & 40 & \baddelta{-0.46 [-1.96, +0.59]} & 1.55 [0.49, 3.38] & \gooddelta{$\times$0.64} & \gooddelta{$\times$0.47} \\
MultiRocket & S=4 & 22 & 13 & 48 & \baddelta{-0.84 [-3.87, +0.65]} & 2.31 [0.79, 4.98] & \gooddelta{$\times$0.49} & \gooddelta{$\times$0.30} \\
\midrule
CNN Classifier & S=2 & 41 & 14 & 18 & \gooddelta{+1.09 [-0.43, +3.00]} & 2.00 [0.87, 5.20] & \gooddelta{$\times$0.92} & \gooddelta{$\times$0.99} \\
CNN Classifier & S=3 & 39 & 11 & 23 & \gooddelta{+0.67 [-1.16, +2.60]} & 2.20 [1.01, 4.27] & \gooddelta{$\times$0.93} & $\times$1.00 \\
CNN Classifier & S=4 & 39 & 14 & 20 & \gooddelta{+1.16 [-0.71, +3.31]} & 2.04 [0.95, 5.87] & \gooddelta{$\times$0.93} & $\times$1.00 \\
\midrule
FCN Classifier & S=2 & 31 & 9 & 33 & \gooddelta{+0.00 [-3.79, +5.55]} & 4.36 [1.76, 12.43] & \gooddelta{$\times$0.79} & \gooddelta{$\times$0.98} \\
FCN Classifier & S=3 & 38 & 4 & 31 & \gooddelta{+0.57 [-4.80, +10.40]} & 6.15 [2.67, 15.76] & \gooddelta{$\times$0.69} & \gooddelta{$\times$0.98} \\
FCN Classifier & S=4 & 30 & 5 & 38 & \baddelta{-1.01 [-8.92, +9.78]} & 9.04 [4.00, 16.88] & \gooddelta{$\times$0.66} & \baddelta{$\times$1.01} \\
\bottomrule
\end{tabular}
}
\endgroup
\end{table}

\begin{table}[t]
\centering
\caption{Cross-backbone consistency of ROMAN's effect on UCR. Spearman rank correlation of the per-dataset accuracy change (\(S=4\) vs.\ \(S=1\)) between backbones, over the 73 datasets evaluated with all four models. \(^{*}\)\(p<0.01\); remaining entries are not significant.}
\label{tab:cross_backbone}
\begin{tabular}{lccc}
\toprule
 & MultiRocket & FCNClassifier & CNNClassifier \\
\midrule
MiniRocket  & \(0.61^{*}\) & \(0.34^{*}\) & \(0.12\) \\
MultiRocket &  & \(0.47^{*}\) & \(-0.13\) \\
FCNClassifier & & & \(-0.04\) \\
\bottomrule
\end{tabular}
\end{table}

\subsection{UEA results}
The UEA benchmark provides a complementary evaluation in the multivariate setting, where ROMAN's pseudochannel construction interacts with existing cross-variable structure. Table~\ref{tab:uea_archive_summary} in \ref{app:tables} reports the same archive-level summaries as for UCR under a common reporting scheme. The picture is qualitatively the same: accuracy changes are mixed and no configuration is significantly different from the baseline (\ref{app:stats}), while inference cost decreases with \(S\) for the ROCKET-family backbones. With only 15 datasets the archive offers little statistical power, which is why the cross-backbone analysis above is carried out on UCR. The multivariate case is also where the channel expansion \(C'=C\sum_s W_s\) is most consequential, since it multiplies an input that is already wide; this is the setting in which the budget-oriented choices of \(\alpha\) and \(S\) discussed in Sect.~\ref{sec:choosing} matter most.

\subsection{Does the multiscale cascade matter?}
\label{sec:cascade}
The synthetic ablation identifies a setting in which the coarser pyramid levels add substantial value for MiniRocket and MultiRocket, beyond the benefit of routing alone (Sect.~\ref{sec:synthetic}). This motivates testing whether the cascade also improves MiniRocket on real datasets. We therefore ablate the cascade from the operator itself: keeping only the finest-scale pseudochannels of ROMAN\((S=4)\), which is the \emph{ROMAN (no multiscale)} condition of the mechanism studies, applied here to the 83 long-sequence UCR datasets with MiniRocket over five seeds. Because the retained windows are the scale-1 block of the very same transform, the two representations share their window geometry exactly and differ only by the coarser scales (\ref{app:cascade}).

The cascade earns its place. The full operator is more accurate than its single-scale ablation on 52 of the 83 datasets against 24 (Wilcoxon signed-rank \(p=0.003\)), and the effect appears on both sides of the distribution: on the ten datasets where ROMAN gains most over the baseline it is better on eight, and on the 40 datasets where ROMAN loses it recovers 1.1 pp of the loss. This is achieved under a handicap, which is what makes the comparison informative: the coarse scales raise the number of pseudochannels from 15 to 26 at the median, so the same fixed budget of 10,000 kernels is spread more thinly over each channel. The cascade therefore has to pay for the dilution it causes and still comes out ahead. Two qualifications keep this in proportion. Restricted to the 23 datasets where ROMAN improves over the baseline, the advantage is directional rather than significant (14 of 23), so the firm statements are the archive-wide one and the reduction of losses. The ablation remains competitive on a minority of tasks. On the two largest gains, CinCECGTorso and GestureMidAirD3, the finest-scale representation alone is sufficient to obtain the improvement, showing that these gains do not require the multiscale cascade.

\section{Using ROMAN in Practice}
\label{sec:practice}

The mechanism studies establish \emph{when} routing should help and the archive benchmarks show how often it does; this section addresses how to use the operator. We first give practical guidance for choosing its two hyperparameters, then present two short case studies outside the archive setting: a very long-sequence task on which the baseline backbone is slow and weak and ROMAN is both more accurate and substantially cheaper, and a simple way to turn the representation diversity induced by different routing scales into an ensemble gain.

\subsection[Choosing S and alpha]{Choosing \(S\) and \(\alpha\)}
\label{sec:choosing}
ROMAN has two hyperparameters. The overlap \(\alpha\) controls how densely each scale is tiled by windows, and hence the redundancy of the representation and the number of pseudochannels. We find that ROMAN is not sensitive to its exact value: on the 83 long-sequence UCR datasets with MiniRocket at \(S=4\), none of \(\alpha\in\{0,0.25,0.75\}\) differs significantly from the \(\alpha=0.5\) used throughout the paper after Holm correction, and \(\alpha=0.25\) and \(\alpha=0.5\) are statistically indistinguishable although the former uses about a quarter fewer pseudochannels (\ref{app:alpha}). In practice \(\alpha\) is therefore a cost knob rather than a quantity to tune: \(\alpha=0.25\) is the natural choice when the number of pseudochannels must be controlled. We use \(\alpha=0.5\) as a simple reference setting that provides approximately double coverage away from the boundaries of scales containing multiple windows.

The number of scales \(S\) is the knob that matters, and it is task-dependent by design. Two rules help. First, \(S\) should keep the base length at the scale of the discriminative structure: the equivalent parameterization \(S^\star \approx 1+\lfloor\log_2(L/L_{\min})\rfloor\) (\ref{app:operator}) makes this explicit, and \(L_{\min}\) should be chosen so that the finest-scale windows still contain the local patterns the task depends on---in the keyword-spotting example below, excessively short windows may truncate discriminative local patterns. Second, because the same dataset can prefer \(S=1\) with one backbone and \(S=4\) with another, \(S\) is best selected on validation data. We measure what such a selection actually yields. For each dataset and seed, the training set is partitioned into four stratified folds; every \(S\in\{1,2,3,4\}\) is fitted on three folds and scored on the fourth; the \(S\) with the highest mean validation accuracy is selected, with ties resolved toward the smaller \(S\); and only the selected configuration is evaluated on the test set. Table~\ref{tab:val_selection} reports this for MiniRocket on the 83 long-sequence UCR datasets (\ref{app:selection}). As a retrospective reference, we select one \(S\) per dataset by maximizing test accuracy averaged across seeds. This estimates the benefit of choosing the best-performing scale among those evaluated for each dataset, averaged over model randomness. We refer to this reference as the dataset-level oracle; because the validation procedure selects \(S\) separately for each dataset and seed, it is not an upper bound on that procedure.

Selection turns a fixed \(S=4\), which loses 2.7 pp on average, into a small but significant gain of 0.8 pp over the \(S=1\) baseline (Wilcoxon signed-rank \(p=0.022\)), achieving 42\% of the dataset-level oracle's average gain over the baseline. Its value lies less in the average than in the tail: a fixed \(S=4\) is worse than the baseline on 40 of the 83 datasets, a validation-selected \(S\) on only 10. This is because the procedure is conservative---it returns \(S=1\) in 55\% of cases, so the median gain is exactly zero---and the honest summary is that it usually leaves the input untouched, occasionally finds a real gain, and rarely does harm. It also has an applicability limit worth stating: on 11 of the 83 datasets the procedure's four-fold feasibility checks fail, and those datasets default to \(S=1\) (they are listed in \ref{app:selection}). Small training sets are exactly where the selection is least reliable and where the identity \(S=1\) remains the sensible default. When no validation data can be spared, the ensemble of Sect.~\ref{sec:diversity} offers another way to benefit from multiple scales without selecting a single configuration. The released library implements both the exact-\(S\) and the \(L_{\min}\)/budget-based selection modes.

\begin{table}[t]
\centering
\caption{Selecting \(S\) on validation data: MiniRocket on the 83 long-sequence UCR datasets, five seeds. Accuracies are averaged over datasets; \(\Delta\) is against the fixed \(S=1\) baseline, with the Wilcoxon signed-rank test over datasets. Wins and losses count datasets differing from the baseline by more than 0.5 pp. The dataset-level oracle selects one \(S\) per dataset by maximizing test accuracy averaged across seeds; it is a retrospective reference, not an upper bound on the validation selector. The final column expresses the average gain over \(S=1\) as a percentage of the oracle's average gain.}
\label{tab:val_selection}
\footnotesize
\setlength{\tabcolsep}{5pt}
\begin{tabular}{lccccc}
\toprule
Selection of \(S\) & Accuracy & \(\Delta\) (pp) & \(p\) & Wins/Losses & Oracle gain \\
\midrule
Fixed \(S=1\) (baseline) & 80.14 & --- & --- & --- & --- \\
Fixed \(S=4\) & 77.45 & \(-2.69\) & 0.017 & 23/40 & --- \\
Validation, four-fold & \(\mathbf{80.90}\) & \(\mathbf{+0.76}\) & \(\mathbf{0.022}\) & 23/10 & 42\% \\
\midrule
Dataset-level oracle & 81.96 & \(+1.81\) & \(1\cdot10^{-9}\) & 39/0 & 100\% \\
\bottomrule
\end{tabular}
\end{table}

\subsection{Very Long Sequences: Keyword Spotting from Raw Audio}
\label{sec:speech}
ROMAN's premise---trading a shorter time axis for more channels---matters most where the time axis is very long. We illustrate this on keyword spotting from raw 16\,kHz audio with the Speech Commands dataset \citep{ref_speechcommands}, using its standard ten-keyword task (\emph{yes, no, up, down, left, right, on, off, stop, go}; \(L=16{,}000\)), with 6,000 training and 1,500 test clips drawn from the official training and test lists. On the raw waveform this is a setting in which ROCKET-style transforms are both slow and weak. Table~\ref{tab:speech} reports the full dependence on \(S\) over three seeds, and shows the knob behaving as designed: accuracy rises steeply from 0.583 on the raw waveform to a plateau of about 0.655 at \(S=2\)--\(3\), then falls back to the baseline by \(S=6\), while cost decreases monotonically throughout. Applying the four-fold validation procedure of Sect.~\ref{sec:choosing} to the training set alone selects \(S=2\), which improves accuracy by 7.1 pp over the raw waveform while at the same time cutting inference from 85 to 67\,ms per clip. The selection lands on the plateau but does not resolve within it: validation separates \(S=2\) from \(S=3\) by 0.25 pp, and on the test set the two differ by 0.2 pp in the opposite direction, so either choice realizes essentially the whole available gain. The decline for \(S\geq5\) is consistent with excessive fragmentation: the finest windows then span at most 1,000 samples, about 60\,ms, which may truncate useful phonetic context. For context, Gwilliams et al. report a mean phoneme duration of 78\,ms (SD 34\,ms) in English narrative speech \citep{ref_phoneme_duration}; this is a reference scale rather than a fixed phoneme duration or a measurement of Speech Commands. Increasing \(S\) also expands the channel dimension and changes how the fixed kernel budget samples the representation, so the experiment does not isolate window duration as the cause. A deployment with a tighter latency budget can read a different operating point off the same curve---\(S=4\) gives up 1.5 pp relative to the selected setting for a further 2.2\(\times\) reduction in inference time---which is precisely the trade-off the operator is meant to expose. The case illustrates both halves of the operator's promise on a task with genuinely long sequences, and the \(L_{\min}\) rule above. For context, specialized end-to-end audio models reach accuracies well above 0.9 on this benchmark; the purpose of the case study is not to compete with them, but to measure what the operator changes for a fixed generic backbone applied directly to the raw waveform.

\begin{table}[t]
\centering
\caption{Keyword spotting from raw audio (Speech Commands, ten-keyword task, \(L=16{,}000\)): MiniRocket on the raw waveform and on ROMAN across the range of \(S\), mean \(\pm\) standard deviation over three seeds. Times are per clip for inference and total for training, single-threaded; the ratio is relative to the \(S=1\) baseline. The configuration selected by four-fold validation on the training set is marked \(\dagger\).}
\label{tab:speech}
\footnotesize
\setlength{\tabcolsep}{4pt}
\begin{tabular}{lccccc}
\toprule
Configuration & Input & Accuracy & Train time & Inference / clip & Ratio \\
\midrule
MiniRocket (\(S=1\)) & \((1,16000)\) & \(0.583 \pm 0.008\) & 559\,s & 85\,ms & 1.00 \\
\(+\) ROMAN(\(S=2\))\(^\dagger\) & \((4,8000)\) & \(\mathbf{0.654 \pm 0.003}\) & 448\,s & 67\,ms & 0.78 \\
\(+\) ROMAN(\(S=3\)) & \((11,4000)\) & \(0.656 \pm 0.008\) & 355\,s & 52\,ms & 0.60 \\
\(+\) ROMAN(\(S=4\)) & \((26,2000)\) & \(0.639 \pm 0.004\) & 228\,s & 30\,ms & 0.35 \\
\(+\) ROMAN(\(S=5\)) & \((57,1000)\) & \(0.592 \pm 0.008\) & 166\,s & 20\,ms & 0.23 \\
\(+\) ROMAN(\(S=6\)) & \((120,500)\) & \(0.576 \pm 0.007\) & 130\,s & 14\,ms & 0.16 \\
\bottomrule
\end{tabular}
\end{table}

\subsection{Leveraging Representation Diversity}
\label{sec:diversity}
As a final example, we test whether the representation diversity induced by different values of \(S\) can be exploited in a practical but easily interpretable way, through a deliberately simple proof-of-concept rather than a new ensemble method. For each backbone and archive, we compare two five-model hard-voting ensembles trained under the same protocol: a baseline-only ensemble composed of five standard models, and a mixed-scale ensemble composed of two standard models together with three ROMAN-based models using \(S=2\), \(S=3\), and \(S=4\). The two ensembles therefore differ only in whether part of the ensemble operates on ROMAN-transformed inputs at different routing scales. An extended description of the setup and results is given in \ref{app:ensemble}.

Table~\ref{tab:optional_ensemble_summary} summarizes the comparison through wins, ties, losses, and the accuracy difference, defined as mixed-scale ensemble accuracy minus baseline-only ensemble accuracy. The results are intended only as an illustrative case study, but they show that a simple mixed-scale construction is often better than the baseline-only ensemble. This supports the narrow practical claim made here: different values of \(S\) can induce complementary representations with complementary prediction behavior, and even a simple ensemble can exploit that diversity.

\begin{table}[t]
\centering
\caption{Illustrative five-model hard-voting ensemble study: Mixed-scale ROMAN ensemble vs. baseline ensemble. Accuracy difference is reported as Median[\(Q_1,Q_3\)].}
\label{tab:optional_ensemble_summary}
\begin{tabular}{llcccc}
\toprule
Archive & Model & Wins & Ties & Losses & Acc. diff. \\
\midrule
UCR & MiniRocket & 37 & 23 & 23 & \gooddelta{0.00 [-0.65, 1.92]} \\
UCR & MultiRocket & 32 & 26 & 25 & \gooddelta{0.00 [-0.80, 1.29]} \\
UCR & FCN & 41 & 10 & 22 & \gooddelta{1.10 [-2.29, 7.47]} \\
UCR & CNN & 42 & 18 & 13 & \gooddelta{1.28 [0.00, 3.59]} \\
\midrule
UEA & MiniRocket & 7 & 2 & 6 & \gooddelta{0.40 [-1.83, 2.37]} \\
UEA & MultiRocket & 6 & 5 & 4 & \baddelta{0.00 [-0.84, 0.72]} \\
UEA & FCN & 6 & 4 & 5 & \gooddelta{0.00 [-0.89, 1.32]} \\
UEA & CNN & 9 & 2 & 4 & \gooddelta{1.37 [-0.73, 3.90]} \\
\bottomrule
\end{tabular}
\end{table}

\section{Discussion}

\paragraph{\textbf{When ROMAN Helps}}
Taken together, the synthetic and archive results support a task-structure interpretation of ROMAN. The operator is most useful when class information depends not only on motif presence, but also on coarse temporal position, on relations between distant events, or on interactions between fine- and coarse-scale structure. In these regimes, routing scale and coarse temporal position into pseudochannels makes that information more accessible to standard convolutional backbones, especially when it would otherwise need to be recovered only indirectly through dilation or after aggressive temporal pooling. The ablations indicate that explicit routing and explicit multiscale views serve complementary roles. Routing alone can make the relevant temporal relationships accessible, while the pyramid adds value when the downstream features do not recover those relationships effectively from the single-scale representation. This distinction is particularly clear on the multiscale-interaction task, where the ROCKET models benefit substantially from the pyramid but the learned CNN and FCN already perform almost perfectly without it. The coordinate-channel results further suggest that routing can make positional information more accessible to the tested ROCKET backbones than simply appending coordinates, whereas FCNClassifier can exploit the coordinate channel directly on the position-awareness task. ROMAN is not a universal improvement: when the task is well described by translation-invariant motif detection, pooled baselines already have an appropriate inductive bias, and ROMAN may reduce performance by weakening a useful invariance.

Our experiments suggest that task structure determines the opportunity for ROMAN to help, while the backbone influences how much of that opportunity is realized. Routing may be useful when classification requires coarse positional awareness, interactions across scales, or relationships between distant, rapidly varying local events. The gain should depend on how readily the backbone already captures that information. Conversely, when class identity is invariant under arbitrary temporal shifts, positional routing offers no obvious advantage and may weaken a useful invariance, as illustrated by the synthetic invariance task.

\paragraph{\textbf{Limitations and Scope}}
The scope of the present study is constrained in three ways. First, the useful range of \(S\) is inherently limited: increasing \(S\) shortens the processed temporal axis, but it also expands the pseudochannel dimension rapidly, so the representation can become overly fragmented and, especially for multivariate inputs, unnecessarily large; the keyword-spotting example shows both the gain and its loss along a single \(S\) axis. Second, the archive experiments evaluate a small fixed grid of configurations rather than per-dataset tuning. This keeps the effect of the operator interpretable, and the sensitivity study shows that the results are not driven by the choice of \(\alpha\), but the reported numbers should be read as evidence about controlled changes in representation rather than about optimized deployment; Sect.~\ref{sec:choosing} gives the corresponding practical guidance. Third, the empirical evidence comes from mechanism-driven synthetic tasks, from long-sequence subsets of the UCR and UEA archives, and from one illustrative long-sequence case study. These settings are appropriate for testing the claims of the paper, but they do not span the full range of real-world time-series applications. Accordingly, the contribution is best understood as a controlled operator-level study of when multiscale routing is useful, together with practical guidance on how to apply it, not as a benchmark-optimized recipe.

\paragraph{\textbf{Future Directions}}
First, ROMAN suggests a simple new source of representation diversity: different routing scales produce complementary views of the same time series, which may be useful in stronger ensemble families such as Arsenal or Detach-Rocket-Ensemble-style methods \citep{ref_detach_ens,ref_hive}. The present paper includes only a deliberately minimal proof-of-concept, so optimized ROMAN-based ensembles are left for future work. Second, because ROMAN is differentiable with respect to the input for fixed routing choices, an interesting future direction is to implement a ROMAN-inspired multiscale routing layer inside learned convolutional architectures. Such a layer could reduce temporal length early while preserving coarse positional and cross-scale structure, potentially reducing the need for very deep temporal stacks or extremely large receptive fields.

\section{Conclusion}

We introduced ROMAN, a deterministic multiscale routing operator for time series classification. By rewriting temporal scale and coarse temporal position as pseudochannels, ROMAN changes the representation presented to standard convolutional backbones, with the number of scales acting as a single knob from the untouched input to increasingly routed representations. Across formal analysis, synthetic mechanism studies with an ingredient-level ablation, archive benchmarks, and practical case studies, the results show that this representation change alters inductive bias in a controlled way, that the synthetic results identify conditions under which routing can help, that archive-level patterns are consistent with this interpretation, and that the operator can be applied with simple guidance.

ROMAN's value lies in its simplicity as an addition to the TSC toolbox, alongside rather than in place of strong classifiers: it requires no learning, can be used as a preprocessing step without modifying the downstream model, and often shortens the temporal axis that the classifier must process. In this sense, ROMAN provides an easy-to-use way to make convolutional models more sensitive to temporal structure when class information depends on how events are organized across time and scale.

\paragraph{\textbf{Code availability}}
The ROMAN operator library, including utilities and example use cases, is available at \href{https://github.com/gon-uri/ROMAN}{github.com/gon-uri/ROMAN}. A separate reproduction repository contains the experiment runners, configurations, and analysis code for the results reported in this paper: \href{https://github.com/gon-uri/ROMAN-paper}{github.com/gon-uri/ROMAN-paper}.

\paragraph{\textbf{Data availability}}
This study uses publicly available data from the UCR and UEA time series classification archives and Speech Commands \citep{ref_ucr,ref_uea,ref_speechcommands}. Dataset selections, synthetic-data generators, saved predictions, and analysis outputs are available at \href{https://github.com/gon-uri/ROMAN-paper}{github.com/gon-uri/ROMAN-paper}.

\section*{Funding}
This research did not receive any specific grant from funding agencies in the public, commercial, or not-for-profit sectors.

\section*{Declaration of competing interests}
The author declares that there are no competing interests relevant to this submission.

\section*{Declaration of generative AI and AI-assisted technologies in the manuscript preparation process}
During the preparation of this work, the author used OpenAI's ChatGPT/Codex to assist with coding, manuscript revision, and LaTeX preparation. After using these tools, the author reviewed and edited the content as needed and takes full responsibility for the content of the published article.

\FloatBarrier
\section*{Acknowledgments}
The author thanks Federico Barone and Yago J. S. Cagnacci for valuable discussions of the ideas presented here and for their thoughtful feedback on the manuscript.

\clearpage
\ROMANAppendix

\section{Operator Details and Additional Derivations}
\label{app:operator}

\subsection{Boundary Handling, Odd Lengths, and Realized Scale Lengths}

Before each downsampling step, an odd-length sequence is extended by repeating its final sample. Filtering with \(h=\tfrac14[1,2,1]\) uses edge replication, after which samples at zero-based even indices are retained. Consequently, the realized lengths satisfy \(L_1=L\) and
\[
L_s=\left\lceil\frac{L_{s-1}}{2}\right\rceil
=\left\lceil\frac{L}{2^{s-1}}\right\rceil.
\]
All subsequent quantities in the operator---\(L_{\mathrm{base}}=L_S\), the window counts \(W_s\), and the start indices \(a_{s,w}\)---are defined from these realized lengths.

For the window extraction step, the start indices
\[
a_{s,w}=
\begin{cases}
1, & W_s=1,\\[3pt]
1+\left\lfloor \dfrac{(w-1)(L_s-L_{\mathrm{base}})}{W_s-1}\right\rfloor, & W_s>1,
\end{cases}
\qquad w=1,\dots,W_s,
\]
spread the windows uniformly over the admissible start range. Consequently, the last window ends at the final valid position of the realized scale-\(s\) sequence, and the actual overlap may differ from the target value \(\alpha\) only through integer rounding. In this sense, the operator is exact once the realized \(L_s\) values are fixed.

\subsection{Equivalent Parameterization by a Minimum Admissible Base Length}

Although the main paper treats \(S\) as the primary hyperparameter, ROMAN also admits an equivalent parameterization in terms of a minimum admissible final length \(L_{\min}\). The idea is to choose the deepest pyramid that still preserves at least that final temporal resolution:
\[
S^\star(L_{\min})
=
\max\{S\geq 1:\ L_S \geq L_{\min}\}.
\]
This does not define a different operator; it only defines a different rule for selecting \(S\). Using the idealized dyadic approximation \(L_S\approx L/2^{S-1}\), this becomes
\[
S^\star(L_{\min})
\approx
1+\left\lfloor \log_2\!\left(\frac{L}{L_{\min}}\right)\right\rfloor.
\]
This alternative view is useful when one wants to guarantee that the coarsest routed representation remains long enough for downstream kernels to operate meaningfully. The current benchmark design in the main paper uses this logic implicitly when restricting attention to sufficiently long datasets and fixing \(S \in \{1,2,3,4\}\). 

\subsection{Backbone-Specific Size and Complexity Implications}

The main paper shows that the ROMAN output has size
\[
|\mathbf{Z}| = C' L_{\mathrm{base}}
= C L_{\mathrm{base}} \sum_{s=1}^{S} W_s,
\]
and that the preprocessing cost is
\[
T_{\mathrm{ROMAN}}=\Theta(CL+|\mathbf{Z}|).
\]
What changes downstream is therefore the input shape: the temporal length is reduced from \(L\) to \(L_{\mathrm{base}}\), while the channel dimension grows from \(C\) to \(C'\).

For MiniRocket and MultiRocket, the dominant cost of the transform stage scales primarily with the processed temporal length. With the number of kernels held fixed, replacing \(L\) by \(L_{\mathrm{base}}\) therefore reduces the temporal part of the convolutional feature-extraction cost from order \(L\) to order \(L_{\mathrm{base}}\). The channel expansion does not change this conclusion qualitatively, but it does reduce random channel coverage when \(C'/C\) becomes large. In practice, this is why ROMAN often accelerates ROCKET-style transforms on long sequences, while excessively large pseudochannel expansions may eventually require more kernels to maintain comparable coverage. This is also the reason the paper restricts attention to moderate values \(S\in\{1,2,3,4\}\). 

For CNNClassifier and FCNClassifier, the increase from \(C\) to \(C'\) affects the first convolutional layer directly, since its cost is proportional to the input-channel dimension. If the first layer has \(F_1\) filters of temporal width \(k_1\), its leading-order convolutional cost changes from
\[
\Theta(F_1 C k_1 L)
\quad\text{to}\quad
\Theta(F_1 C' k_1 L_{\mathrm{base}}).
\]
After that first layer, however, subsequent layers operate on learned feature maps rather than on the original \(C'\) pseudochannels. Their cost therefore benefits primarily from the reduction in temporal length from \(L\) to \(L_{\mathrm{base}}\). For FCNClassifier, the final global pooling stage also scales with the shortened temporal axis. For CNNClassifier, the flattening step and subsequent classifier head likewise operate on a shorter temporal representation. This matches the empirical pattern reported in the main paper: ROMAN changes the first layer most directly, while the shorter temporal axis benefits the rest of the network stack. 

In summary, for MiniRocket and MultiRocket, whose transform cost is driven primarily by sequence length, the shorter axis usually dominates, although very large channel expansions can require more kernels to preserve adequate random channel coverage; for CNNClassifier and FCNClassifier, the expanded channel dimension affects only the first convolutional layer directly, whereas the shorter temporal axis benefits the remaining temporal stack. Taken together, these derivations explain the practical efficiency pattern observed in the archive results. ROMAN is not computationally free, because it expands the channel axis, but it often exchanges a moderate channel increase for a more valuable reduction in processed temporal extent. This trade-off is especially favorable on long sequences, which is why the empirical study focuses on long-sequence subsets of the UCR and UEA archives. 
\section[Shared Experimental Environment and Reporting Conventions]{Shared Experimental Environment\\and Reporting Conventions}
\label{app:environment}

\subsection{Hardware and software environment}

All reported experiments for MiniRocket, MultiRocket, CNNClassifier, and FCNClassifier were run on a single MacBook Pro laptop (model identifier \texttt{MacBookPro18,1}) equipped with an Apple M1 Pro chip, a 10-core CPU (8 performance cores and 2 efficiency cores), and 32\,GB of unified memory, running macOS 14.3.1. The experiments use the \texttt{sktime} implementations and wrappers of these four backbones (\texttt{sktime} 0.30.0 for the archive benchmarks and ensemble study; 0.40.1 for the synthetic ablation, the \(\alpha\) sensitivity study, and the keyword-spotting case study, which were run later in the same hardware environment). This information is mainly relevant for the efficiency analysis: the timing values reported in the main paper should be interpreted as comparative measurements obtained in a single fixed environment, rather than as hardware-independent latency claims.

\subsection{Fixed backbone configurations}

For the archive benchmarks, CNNClassifier and FCNClassifier use Adam with learning rate 0.01, batch size 16, and a maximum of 2,000 epochs. Early stopping monitors loss on a stratified 20\% training holdout, restores the best weights, and uses patience 30 for CNNClassifier and 100 for FCNClassifier. The synthetic experiments use patience 100 for both. MiniRocket uses \texttt{num\_kernels}=10,000 in both archives and in the synthetic experiments. MultiRocket uses 6,250 in both archives and in the synthetic experiments. The ROCKET transforms are followed by \texttt{RidgeClassifierCV} with ten logarithmically spaced regularization values from \(10^{-3}\) to \(10^3\).

Package versions are reported above. Process-worker counts and library thread limits are distinct execution settings; configured thread limits describe the requested execution, while effective BLAS thread counts were not independently verified for the historical runs.

\subsection{Shared ROMAN settings}

Across the archive benchmarks, the overlap parameter is fixed to \(\alpha = 0.5\), and the evaluated ROMAN scale settings are
\[
S \in \{1,2,3,4\}.
\]
Throughout the paper, \(S=1\) is the identity and serves as the baseline.

\subsection{Accuracy summaries and practical tie criterion}

In the archive tables, each ROMAN configuration is compared against the corresponding \(S=1\) baseline of the same backbone and archive. Accuracy differences are first averaged across seeds at dataset level, and the resulting per-dataset mean differences are then summarized at archive level.

The main paper reports wins, ties, losses, the median signed accuracy difference, and the median absolute accuracy difference, with the latter two summarized as \(\mathrm{median}\,[Q_1,Q_3]\). A dataset is counted as a tie when the mean accuracy difference across seeds satisfies
\[
|\Delta \mathrm{Acc}| \le 0.005,
\]
that is, when the difference remains within \(0.5\) percentage points of the baseline. This threshold is used as a practical equivalence margin rather than as a formal significance criterion.

\subsection{Timing conventions}

The benchmark runners record three timing components: ROMAN preprocessing time, model fitting time, and post-training prediction time. For paper reporting, these are combined into two phase metrics:
\[
T_{\mathrm{train}} = T_{\mathrm{ROMAN}} + T_{\mathrm{fit}},
\qquad
T_{\mathrm{infer}} = T_{\mathrm{ROMAN}} + T_{\mathrm{predict}}.
\]

For MiniRocket, the \(S=1\) archive baseline uses raw inputs and bypasses ROMAN preprocessing; routed configurations use training-fitted channel normalization. The other archive runners use the normalized identity wrapper at \(S=1\). For the ROCKET-based archive runners, post-training prediction time covers test feature extraction and ridge predictions on the training and test sets, reusing training features computed during fitting.

The training-time and inference-time ratios reported in the archive tables are computed by: (i) averaging each phase time across seeds on each dataset, (ii) dividing that mean by the corresponding baseline mean on the same dataset, and (iii) summarizing the resulting per-dataset ratios by their median across datasets.

\section{Synthetic Dataset Construction}
\label{app:synthetic}

\subsection{Common synthetic protocol}
The synthetic experiments are designed as mechanism studies rather than benchmark optimization exercises. Across all reported synthetic tasks, the shared protocol is as follows:
\begin{itemize}
\item balanced binary classification;
\item sequence length fixed by task family;
\item 250 test instances per realization, and 500 training instances per realization for the position-awareness, multiscale-interaction, and positional-invariance tasks; the long-range correlation task uses 1000 training instances per realization;
\item 10 independent realizations per reported comparison, with the same realization seeds shared by all input representations and backbones;
\item per-series \(z\)-normalization;
\item fixed ROMAN overlap parameter \(\alpha = 0.5\);
\item comparison, with the same downstream backbone, of the four input representations described in Sect.~\ref{app:ablation}: raw input (Base), raw input with a coordinate channel (Coord), ROMAN without its multiscale cascade, and \(\mathrm{ROMAN}(S=4)\).
\end{itemize}

\subsection{Position-awareness dataset}

The position-awareness dataset uses univariate series of length 512. Each series contains exactly two identical single-sample spikes embedded in Gaussian background noise with standard deviation 1.0. If the left spike is placed at distance \(d\) from the left boundary, the right spike is placed symmetrically at the same distance from the right boundary.

The experimental configuration is:
\begin{itemize}
\item \(d_{\min}=32\),
\item center safety margin \(s_{\min}=128\),
\item inter-class gap parameter \(=16\),
\item spike amplitude sampled uniformly from \([4.0, 4.5]\).
\end{itemize}

The admissible distance interval is partitioned into two disjoint class ranges, so that the label depends only on the coarse distance regime and not on motif identity. Because the two spikes are identical and symmetric, the task isolates coarse positional information while avoiding simple asymmetric shortcuts.

\subsection{Long-range correlation dataset}

The long-range correlation dataset uses univariate series of length 512 with two fixed sparse burst windows centered near \(L/6\) and \(5L/6\). Each burst occupies a local window of length 33 and contains 4 spikes with minimum within-burst spacing 8.

The experimental configuration is:
\begin{itemize}
\item burst length \(=33\),
\item 4 spikes per burst,
\item minimum within-burst spacing \(=8\),
\item minimum pattern distance \(=6\),
\item pattern-library size \(=4\),
\item spike amplitude \(=5.0\),
\item Gaussian noise standard deviation \(=1.0\),
\item 1000 training instances per realization.
\end{itemize}

The class label depends on the relation between the two distant bursts:
\begin{itemize}
\item class 0: the two bursts use the same local pattern;
\item class 1: the two bursts use different local patterns.
\end{itemize}

Because the burst geometry is fixed across classes, the task cannot be solved from coarse position alone. Its purpose is to test whether a model can compare distant local structures rather than merely detect their presence.

\subsection{Multiscale interaction dataset}

The multiscale interaction dataset uses univariate series of length 512 built from a full-length low-frequency cosine and a localized high-frequency cosine burst. The local burst has width 32 and is embedded inside a wider masked region of width 96, so that the coarse phase is not directly visible within the local support of the burst.

The experimental configuration is:
\begin{itemize}
\item 1 coarse cosine cycle across the full series,
\item burst length \(=32\),
\item 4 fine oscillation cycles within the burst,
\item mask length \(=96\),
\item 4 discrete phase values,
\item fine amplitude \(=4.0\),
\item coarse amplitude \(=2.0\),
\item Gaussian noise standard deviation \(=0.75\).
\end{itemize}

The class label depends on whether the fine burst phase matches the coarse phase implied by the outer context:
\begin{itemize}
\item class 0: coarse and fine phase agree;
\item class 1: coarse and fine phase differ.
\end{itemize}

This construction is designed to test cross-scale binding rather than simple local detection.

\subsection{Full positional invariance dataset}

The full positional invariance dataset serves as the negative-control synthetic task. It uses univariate series of length 512 containing a small number of local sparse motifs placed at random valid positions over the full timeline.

The experimental configuration is:
\begin{itemize}
\item motif length \(=33\),
\item 4 spikes per motif,
\item minimum within-motif spacing \(=3\),
\item 2 distractor patterns in the motif library,
\item 1 distractor instance per series,
\item 1 target instance in each positive example,
\item minimum target--distractor pattern distance \(=6\),
\item minimum separation between motif instances \(=16\),
\item motif amplitude \(=6.0\),
\item Gaussian noise standard deviation \(=1.0\).
\end{itemize}

Positive and negative examples contain the same total number of motif instances. In the positive class, one distractor slot is replaced by the target motif; in the negative class, all instances are distractors. Motif positions are sampled i.i.d. over the full valid start range in both training and test data. The dataset is therefore intentionally aligned with translation-invariant motif detection and serves as a negative control for ROMAN.

\subsection{Input representations compared in the mechanism studies}
\label{app:ablation}

Each synthetic task is evaluated, with the same downstream backbone and the same realization seeds, under four input representations. Their purpose is to attribute the effects reported in the main text to specific ingredients of the operator.

\begin{itemize}
\item \textbf{Base}: the raw series, of shape \((1,512)\).
\item \textbf{Coord}: the raw series with one additional channel holding a linear ramp from 0 to 1 over the time axis, of shape \((2,512)\). This is the one-dimensional analogue of the CoordConv input of Liu et al., and injects absolute position without restructuring the signal or shortening the time axis.
\item \textbf{ROMAN without multiscale}: ROMAN without its multiscale cascade, i.e., the finest-scale windows of \(\mathrm{ROMAN}(S=4)\) only --- the first \(W_1\) pseudochannels of the ROMAN output, of shape \((W_1, L_{\mathrm{base}}) = (15, 64)\) for \(L=512\) and \(\alpha=0.5\). It routes coarse position into channels and shortens the time axis exactly as ROMAN does, but contains no coarser scale. Because it is literally a slice of the ROMAN output, the two conditions share window geometry by construction.
\item \textbf{ROMAN}: the full operator with \(S=4\) and \(\alpha=0.5\), of shape \((26,64)\).
\end{itemize}

\paragraph{Marking conventions} In the main-text synthetic tables, bold and underlining mark the highest and second-highest distinct mean accuracies, respectively, at the displayed precision in each row; ties receive the same marking. These rankings are descriptive and do not indicate statistical equivalence. Gray marks accuracies that do not exceed the independent random-guessing reference at the 5\% level. For each cell, we pool the number of correct test predictions across the ten realizations (\(10\times 250 = 2500\) predictions) and compute the upper-tail probability under \(K\sim\mathrm{Binomial}(2500,0.50)\). Entries with a one-sided \(p\)-value of at least 0.05 are gray; these values are not corrected for multiple comparisons. This is a descriptive comparison with independent, equiprobable random guesses against the fixed test labels, not a class-count-preserving label-permutation test or a test of consistency across realizations.

\paragraph{Paired differences} Table~\ref{tab:ablation_deltas} reports, for each condition, the paired accuracy difference with respect to Base (mean and standard deviation over the ten realizations, in percentage points), which complements the absolute accuracies given in the main text.

\begin{table}[t]
\centering
\caption{Paired accuracy differences with respect to the raw-input baseline (percentage points, mean \(\pm\) standard deviation over ten realizations) for the three alternative input representations, per synthetic task and backbone.}
\label{tab:ablation_deltas}
\footnotesize
\setlength{\tabcolsep}{5pt}
\begin{tabular}{llrrr}
\toprule
Task & Backbone & Coord & ROMAN & ROMAN \\
 & & & (no multiscale) & (\(S{=}4\)) \\
\midrule
\multirow{4}{*}{Position awareness}
 & MiniRocket & \(+1.0 \pm 2.7\) & \(+19.3 \pm 3.1\) & \(+19.8 \pm 3.2\) \\
 & MultiRocket & \(-2.3 \pm 2.9\) & \(+17.4 \pm 4.5\) & \(+16.5 \pm 3.2\) \\
 & CNNClassifier & \(-2.0 \pm 9.2\) & \(-26.4 \pm 3.5\) & \(-26.2 \pm 3.4\) \\
 & FCNClassifier & \(+49.1 \pm 2.0\) & \(+32.8 \pm 4.4\) & \(+28.4 \pm 4.2\) \\
\midrule
\multirow{4}{*}{Long-range correlation}
 & MiniRocket & \(+0.2 \pm 2.0\) & \(+19.8 \pm 6.0\) & \(+16.2 \pm 2.2\) \\
 & MultiRocket & \(-1.7 \pm 2.6\) & \(+16.8 \pm 6.5\) & \(+12.4 \pm 8.0\) \\
 & CNNClassifier & \(+0.5 \pm 2.7\) & \(+39.0 \pm 5.9\) & \(+40.3 \pm 5.3\) \\
 & FCNClassifier & \(-0.6 \pm 2.2\) & \(+50.4 \pm 3.2\) & \(+50.3 \pm 2.9\) \\
\midrule
\multirow{4}{*}{Multiscale interaction}
 & MiniRocket & \(-0.8 \pm 2.3\) & \(+3.7 \pm 4.6\) & \(+18.8 \pm 7.1\) \\
 & MultiRocket & \(+0.3 \pm 3.3\) & \(+4.5 \pm 5.9\) & \(+20.7 \pm 6.2\) \\
 & CNNClassifier & \(-0.1 \pm 1.7\) & \(+49.0 \pm 2.8\) & \(+49.2 \pm 3.1\) \\
 & FCNClassifier & \(-0.4 \pm 4.1\) & \(+50.0 \pm 2.7\) & \(+50.0 \pm 2.7\) \\
\midrule
\multirow{4}{*}{Positional invariance}
 & MiniRocket & \(-0.2 \pm 1.3\) & \(-28.5 \pm 3.0\) & \(-31.3 \pm 4.2\) \\
 & MultiRocket & \(-6.0 \pm 5.1\) & \(-21.4 \pm 4.0\) & \(-20.7 \pm 6.4\) \\
 & CNNClassifier & \(+1.1 \pm 2.7\) & \(-0.7 \pm 4.4\) & \(-1.1 \pm 3.2\) \\
 & FCNClassifier & \(-0.0 \pm 0.1\) & \(-48.8 \pm 5.5\) & \(-47.9 \pm 3.7\) \\
\bottomrule
\end{tabular}
\end{table}

\section{Archive Benchmark Protocol Details}
\label{app:archive}

\subsection{Archive inclusion rule}

For both UCR and UEA, the archive benchmark analysis is restricted to datasets whose time-series length satisfies
\[
L \ge 256.
\]
This filter ensures that nontrivial ROMAN configurations remain meaningful and that the coarsest routed representation is not dominated by excessive downsampling.

\subsection{Additional exclusion rule for the learned convolutional models}

CNNClassifier and FCNClassifier are trained with an early-stopping protocol that requires an internal validation split. For these two models only, datasets are excluded when some classes are too underrepresented in the training set for that split to remain well posed.

These exclusions are protocol-driven rather than performance-driven. The issue is not low test accuracy, but the inability of the early-stopping wrapper to construct a representative validation subset for those class distributions.

\subsection{Explicit UCR exclusion list for CNN and FCN}

Among the long-sequence UCR candidates, the datasets excluded from the CNN and FCN presentation are:
\begin{itemize}
\item \texttt{DiatomSizeReduction}
\item \texttt{FaceFour}
\item \texttt{FiftyWords}
\item \texttt{Mallat}
\item \texttt{Phoneme}
\item \texttt{PigAirwayPressure}
\item \texttt{PigArtPressure}
\item \texttt{PigCVP}
\item \texttt{Symbols}
\item \texttt{WordSynonyms}
\end{itemize}

The filter \(L \ge 256\) yields 83 UCR candidate datasets. After applying the validation-split feasibility rule above, 73 datasets remain in the UCR CNN and FCN analysis.

\section{Statistical Tests on the Archive Benchmarks}
\label{app:stats}

This section reports, for completeness, formal paired tests of each ROMAN configuration against the corresponding \(S=1\) baseline, computed on the per-dataset mean accuracies over the five seeds. For every backbone--archive pair we apply a two-sided Wilcoxon signed-rank test to the paired accuracy differences and correct the resulting \(p\)-values with the Holm procedure within the family \(S\in\{2,3,4\}\). As an effect size we report the Hodges--Lehmann estimate of the paired difference (HL~\(\Delta\), in percentage points). Table~\ref{tab:stat_tests} lists the results. Six of the 24 comparisons are significant after correction: CNNClassifier is significantly better with ROMAN at every \(S\) on UCR (HL~\(\Delta\approx+1\) pp), and MultiRocket is significantly worse at every \(S\) on UCR, with a deficit that grows with \(S\). All remaining comparisons, including every comparison on UEA (\(n=15\)), are statistically indistinguishable from the baseline. As discussed in the main text, these aggregate tests average over the task-dependent heterogeneity that ROMAN is designed to expose; the per-dataset and cross-backbone analyses in the main text are the more informative view.

\begin{figure}[t]
\centering
\includegraphics[width=0.48\textwidth]{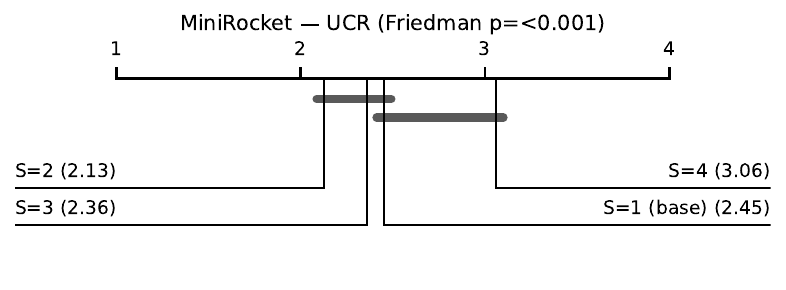}\hfill
\includegraphics[width=0.48\textwidth]{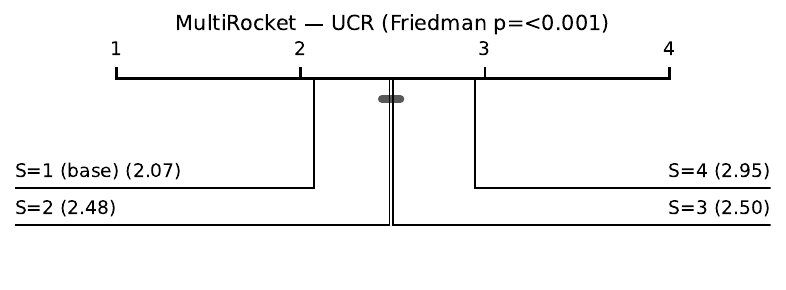}\\[2mm]
\includegraphics[width=0.48\textwidth]{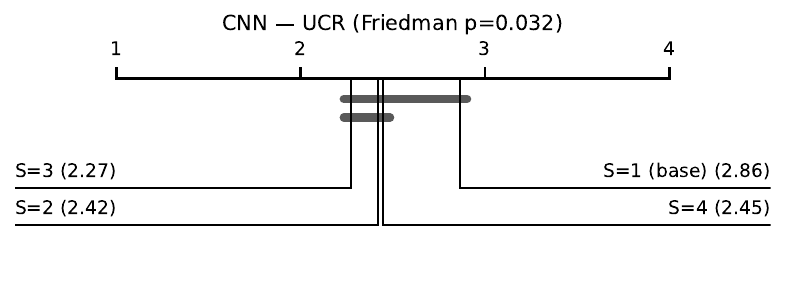}\hfill
\includegraphics[width=0.48\textwidth]{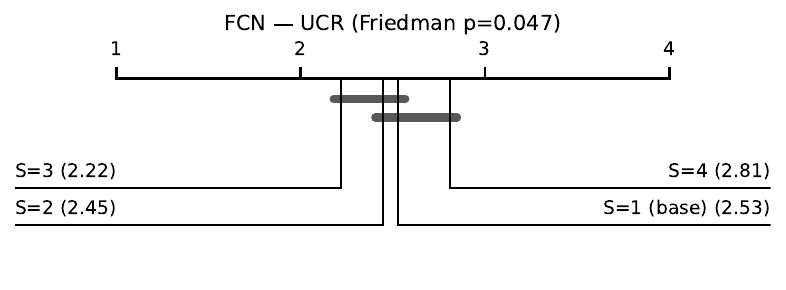}
\caption{Critical-difference diagrams over the ROMAN configurations \(S\in\{1,2,3,4\}\) for each backbone on the long-sequence UCR benchmark (average ranks across datasets; lower is better). Configurations connected by a horizontal bar are not significantly different (pairwise Wilcoxon signed-rank tests with Holm correction, \(\alpha=0.05\)).}
\label{fig:cd_single}
\end{figure}

Figure~\ref{fig:cd_single} complements the paired tests with critical-difference diagrams over the four configurations \(S\in\{1,2,3,4\}\) for each backbone on UCR (Friedman test followed by pairwise Wilcoxon--Holm comparisons; models joined by a bar are not significantly different). The Friedman test is significant for all four UCR backbones; the best average rank is obtained by \(S=2\) for MiniRocket, \(S=3\) for CNNClassifier and FCNClassifier, and \(S=1\) for MultiRocket. On UEA no Friedman test reaches significance except for FCNClassifier (\(p=0.044\)), consistent with the small number of datasets.

\begin{table}[t]
\centering
\caption{Paired tests per backbone and archive: each ROMAN configuration vs.\ the \(S=1\) baseline on per-dataset mean accuracies. W/T/L are wins, ties (\(|\Delta|\le 0.5\) pp), and losses. HL~\(\Delta\) is the Hodges--Lehmann pseudo-median of the paired accuracy difference in percentage points. \(p\)-values are two-sided Wilcoxon signed-rank, Holm-corrected within each backbone--archive family; bold marks \(p<0.05\) after correction.}
\label{tab:stat_tests}
\begin{tabular}{llcccrc}
\toprule
Archive & Model & \(S\) & \(n\) & W/T/L & HL~\(\Delta\) (pp) & \(p\) (Holm) \\
\midrule
UCR & MiniRocket & 2 & 83 & 35/22/26 & \(+0.31\) & 0.504 \\
UCR & MiniRocket & 3 & 83 & 32/18/33 & \(-0.02\) & 0.933 \\
UCR & MiniRocket & 4 & 83 & 23/20/40 & \(-0.94\) & 0.050 \\
UCR & MultiRocket & 2 & 83 & 23/23/37 & \(-0.48\) & \textbf{0.019} \\
UCR & MultiRocket & 3 & 83 & 22/21/40 & \(-0.77\) & \textbf{0.017} \\
UCR & MultiRocket & 4 & 83 & 22/13/48 & \(-1.39\) & \textbf{0.005} \\
UCR & CNNClassifier & 2 & 73 & 41/14/18 & \(+1.09\) & \textbf{0.022} \\
UCR & CNNClassifier & 3 & 73 & 39/11/23 & \(+0.95\) & \textbf{0.039} \\
UCR & CNNClassifier & 4 & 73 & 39/14/20 & \(+1.00\) & \textbf{0.039} \\
UCR & FCNClassifier & 2 & 73 & 31/9/33 & \(+0.81\) & 0.848 \\
UCR & FCNClassifier & 3 & 73 & 38/4/31 & \(+1.36\) & 0.848 \\
UCR & FCNClassifier & 4 & 73 & 30/5/38 & \(+0.13\) & 0.891 \\
\midrule
UEA & MiniRocket & 2 & 15 & 6/1/8 & \(-0.60\) & 0.991 \\
UEA & MiniRocket & 3 & 15 & 10/1/4 & \(+0.92\) & 0.991 \\
UEA & MiniRocket & 4 & 15 & 7/4/4 & \(+0.49\) & 0.991 \\
UEA & MultiRocket & 2 & 15 & 5/5/5 & \(-0.12\) & 1.000 \\
UEA & MultiRocket & 3 & 15 & 5/4/6 & \(-0.27\) & 1.000 \\
UEA & MultiRocket & 4 & 15 & 5/0/10 & \(-0.74\) & 1.000 \\
UEA & CNNClassifier & 2 & 15 & 5/3/7 & \(-0.03\) & 1.000 \\
UEA & CNNClassifier & 3 & 15 & 7/1/7 & \(-0.04\) & 1.000 \\
UEA & CNNClassifier & 4 & 15 & 7/1/7 & \(+0.03\) & 1.000 \\
UEA & FCNClassifier & 2 & 15 & 10/1/4 & \(+1.13\) & 0.250 \\
UEA & FCNClassifier & 3 & 15 & 8/1/6 & \(+0.85\) & 0.977 \\
UEA & FCNClassifier & 4 & 15 & 5/2/8 & \(-0.82\) & 0.977 \\
\bottomrule
\end{tabular}
\end{table}

\section[Sensitivity to the Overlap Parameter alpha]{Sensitivity to the Overlap Parameter \(\alpha\)}
\label{app:alpha}

All experiments in the paper use \(\alpha=0.5\). To assess the sensitivity of the results to this choice we re-ran MiniRocket on the full long-sequence UCR population (83 datasets, five seeds, the protocol of the main paper) with ROMAN at \(S=4\) and \(\alpha\in\{0,0.25,0.5,0.75\}\), together with the \(S=1\) baseline. Table~\ref{tab:alpha} summarizes the result. Relative to the \(\alpha=0.5\) setting, the paired per-dataset accuracy difference is \(-0.37\) pp at \(\alpha=0\), \(+0.01\) pp at \(\alpha=0.25\), and \(-0.26\) pp at \(\alpha=0.75\); none is significant after Holm correction within this three-test family (the raw \(p=0.020\) at \(\alpha=0.75\) becomes \(0.06\)). The sign of ROMAN's effect is stable across \(\alpha\) on every dataset with a large effect (the largest gains and the largest losses at \(\alpha=0.5\) remain gains and losses at every other \(\alpha\)), so \(\alpha\) trades redundancy and pseudochannel count against small accuracy changes without altering which datasets benefit. Because the finest scale contributes most of the windows, the pseudochannel count grows less than the idealized \(1/(1-\alpha)\): relative to \(\alpha=0.5\), the median count is \(0.58\times\) at \(\alpha=0\), \(0.77\times\) at \(\alpha=0.25\), and \(1.85\times\) at \(\alpha=0.75\).

\begin{table}[t]
\centering
\caption{Sensitivity to \(\alpha\): MiniRocket with ROMAN at \(S=4\) on the 83 long-sequence UCR datasets (five seeds). Mean accuracy over datasets; paired difference vs.\ the \(S=1\) baseline (mean and median over datasets, in pp; W/L = datasets with \(|\Delta|>0.5\) pp); paired difference vs.\ the \(\alpha=0.5\) setting with two-sided Wilcoxon \(p\)-values, raw and Holm-corrected; and the median pseudochannel count relative to \(\alpha=0.5\).}
\label{tab:alpha}
\footnotesize
\setlength{\tabcolsep}{3.5pt}
\begin{tabular}{ccccccccc}
\toprule
 & & \multicolumn{3}{c}{vs.\ \(S{=}1\) (pp)} & \multicolumn{3}{c}{vs.\ \(\alpha{=}0.5\) (pp)} & \\
\cmidrule(lr){3-5}\cmidrule(lr){6-8}
\(\alpha\) & Mean acc.\ (\%) & mean & median & W / L & \(\Delta\) & \(p\) raw & \(p\) Holm & Rel.\ ch. \\
\midrule
0    & 77.08 & \(-3.06\) & \(-0.92\) & 20 / 46 & \(-0.37\) & 0.147 & 0.294 & 0.58 \\
0.25 & 77.46 & \(-2.69\) & \(-0.40\) & 27 / 41 & \(+0.01\) & 0.820 & 0.820 & 0.77 \\
0.5  & 77.45 & \(-2.69\) & \(-0.44\) & 23 / 40 & --- & --- & --- & 1.00 \\
0.75 & 77.19 & \(-2.96\) & \(-0.85\) & 25 / 46 & \(-0.26\) & 0.020 & 0.060 & 1.85 \\
\bottomrule
\end{tabular}
\end{table}

\section[Keyword Spotting from Raw Audio: Protocol and Dependence on S]{Keyword Spotting from Raw Audio: Protocol and Dependence on \(S\)}
\label{app:speech}

The keyword-spotting case study of the main paper uses the Speech Commands dataset (v0.02) restricted to its standard ten-keyword task. Clips are one-second 16\,kHz waveforms (\(L=16{,}000\); shorter clips are zero-padded), scaled to \([-1,1]\) and otherwise used raw. We draw 600 training clips per keyword from the official training list (6,000 in total) and 150 test clips per keyword from the official test list (1,500), with a fixed random subsample. The backbone is MiniRocket with 10,000 kernels followed by a ridge classifier, exactly as in the archive benchmarks; ROMAN uses \(\alpha=0.5\). Three seeds control MiniRocket's random kernels; the data subsample is fixed. Timings are single-threaded and reported per clip for inference (ROMAN transform plus MiniRocket transform plus classification) and in total for training.

The dependence on \(S\) is reported in the keyword-spotting table of the main text. At \(S=5\) the base length is 1,000 samples (about 60\,ms) and at \(S=6\) it is 500 (about 30\,ms), shorter than typical phonetic units, while the pseudochannel count reaches 57 and 120; both effects remove the gain, which is why accuracy returns to the baseline level at the deepest settings.

The value of \(S\) reported in the main text is selected by the four-fold procedure described there, applied to the 6,000 training clips only: the training set is partitioned into four stratified folds, each \(S\in\{1,\dots,6\}\) is fitted on three folds and scored on the fourth, and the \(S\) with the highest mean validation accuracy is selected, with ties resolved toward the smaller \(S\). The test clips play no part in this choice. The mean validation accuracies are 0.571, 0.657, 0.654, 0.637, 0.593 and 0.572 for \(S=1,\dots,6\), so the procedure selects \(S=2\). They track the test accuracies closely in shape while sitting slightly lower, as expected when each fold is fitted on three quarters of the training clips, and they reproduce the plateau at \(S=2\)--\(3\) and the decline beyond it. The margin between \(S=2\) and \(S=3\) is 0.25 pp on validation and 0.2 pp on test in the opposite direction, which is well inside the noise of either estimate: the procedure identifies the plateau reliably but should not be read as resolving the ordering within it.

\section[Selecting S on Validation Data: Protocol and Coverage]{Selecting \(S\) on Validation Data: Protocol and Coverage}
\label{app:selection}

The validation-selection results of the main text are produced as follows. For each of the 83 long-sequence UCR datasets and each of the five seeds, the training set is partitioned into four stratified folds. Every \(S\in\{1,2,3,4\}\) is fitted on three folds and scored on the held-out fold, and the \(S\) with the highest mean validation accuracy is selected, ties being resolved toward the smaller \(S\). Only the selected configuration is then evaluated on the test set, with the operator and backbone settings used throughout the paper (\(\alpha=0.5\), \(L_{\min}\) constraint of 32 timesteps per pseudochannel, 10,000 kernels, ridge classifier). The seed controls both the partition and the random kernels, so each seed performs its own selection.

Validation selection is attempted only when the constructed four-fold partition contains every class in every validation fold. This is a conservative coverage requirement, not a condition for computing accuracy, and requires at least four examples per class. The procedure defaults to \(S=1\) on the 11 datasets that fail its feasibility checks. These are DiatomSizeReduction, FiftyWords, Phoneme, PigAirwayPressure, PigArtPressure, PigCVP, Mallat, WordSynonyms, FaceFour, InsectEPGSmallTrain, and Symbols.

\section{Ablating the Multiscale Cascade}
\label{app:cascade}

The ablation of the multiscale cascade reported in the main text removes the coarse scales from the operator while leaving everything else identical. Concretely, ROMAN is fitted at \(S=4\) as usual and only the scale-1 block of pseudochannels is retained, so the windows that survive are exactly those the full operator places at the finest scale; the number of such windows follows from the series length and is not fixed in advance (15 of 26 pseudochannels at the median across the archive). The comparison therefore isolates the contribution of the cascade: the two representations share their window geometry, their overlap and their base length, and differ only in whether the coarser scales are present. The backbone, kernel budget, seeds and evaluation protocol are those of the archive benchmarks, and the baseline and full-ROMAN columns are the same runs reported there.

\section{Ensemble Case Study Protocol}
\label{app:ensemble}

\begin{figure}[t]
\centering
\includegraphics[width=0.9\linewidth]{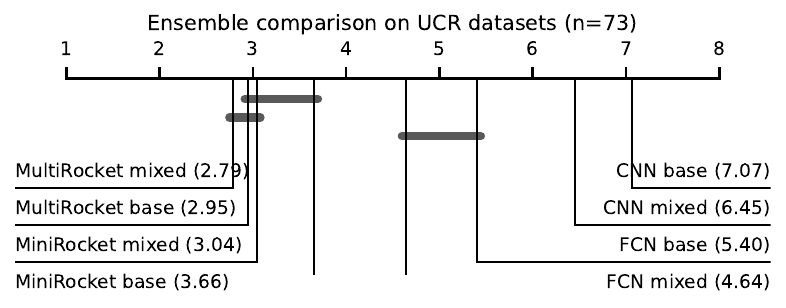}\par
\vspace{0.5em}
\includegraphics[width=0.9\linewidth]{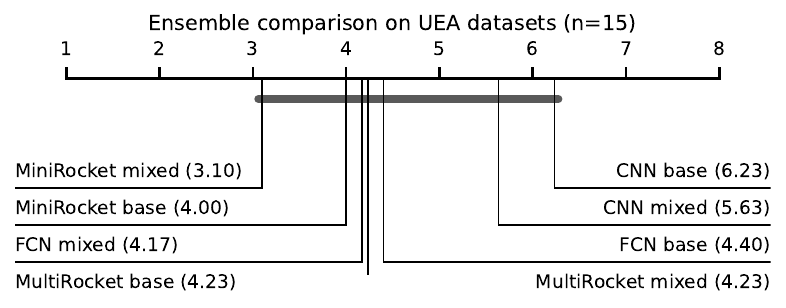}
\caption{Critical-difference diagrams for the five-model hard-voting ensemble study on UCR (top, 73 shared datasets) and UEA (bottom, 15). Each diagram compares the eight ensemble variants within the archive: the four baseline-only ensembles and their four mixed-scale ROMAN counterparts; lower average rank is better. Variants connected by a horizontal bar are not significantly different (Friedman test followed by pairwise Wilcoxon signed-rank tests with Holm correction, \(\alpha=0.05\); the same procedure as Fig.~\ref{fig:cd_single}). On UEA the Friedman test is significant but no pairwise difference survives correction, so all eight variants form one group.}
\label{fig:optional_ensemble_cd}
\end{figure}

\subsection{Ensemble construction}

The main paper reports a five-model hard-voting protocol. For each backbone and archive, it compares:
\begin{itemize}
\item a baseline-only ensemble built from seeds \(0,1,2,3,4\) of the \(S=1\) model;
\item a mixed-scale ensemble built from baseline seeds \(0,1\) together with seed 0 of \(S=2\), seed 0 of \(S=3\), and seed 0 of \(S=4\).
\end{itemize}

This construction was chosen so that both ensembles contain the same number of members, use the same training data, and differ only in whether some members receive ROMAN preprocessing at different routing scales.

\subsection{Voting and reported quantities}

The ensemble result reported in the main paper uses hard voting. The reported difference is always
\[
\Delta \mathrm{Acc}
=
\mathrm{Acc}_{\mathrm{mixed\mbox{-}scale}}
-
\mathrm{Acc}_{\mathrm{baseline\mbox{-}only}}.
\]

For each model--archive pair, the compact ensemble table reports wins, ties, losses, and the median signed ensemble accuracy difference, summarized as \(\mathrm{median}\,[Q_1,Q_3]\). The same practical tie criterion used in the archive tables is applied here: a dataset is counted as a tie when the corresponding mean accuracy difference remains within \(0.5\) percentage points of zero. The archive-level comparison is complemented by the critical-difference diagrams in Fig.~\ref{fig:optional_ensemble_cd}, which summarize average ranks across shared datasets for the eight ensemble variants considered in each archive, using the same test procedure as the single-model diagrams of \ref{app:stats}. On UCR, each backbone's mixed-scale ensemble ranks above its baseline-only counterpart, with MultiRocket and MiniRocket forming the top group; on UEA the ranking is the same in direction but no pairwise difference is significant.

\section{Complete Archive Result Tables}
\label{app:tables}

This section presents the complete archive results for UCR and UEA. Table~\ref{tab:uea_archive_summary} gives the archive-level UEA summary in the same format as the UCR summary of the main paper. The subsequent tables give the per-dataset results for UCR and UEA.

\begin{table}[!htb]
\centering
\caption{Archive-level summary of ROMAN on the long-sequence UEA benchmark.}
\label{tab:uea_archive_summary}
\begingroup
\setlength{\tabcolsep}{3.5pt}
\resizebox{\textwidth}{!}{%
\begin{tabular}{llccccccc}
\toprule
Model & Config. & Wins & Ties & Losses & Acc. diff. & Absolute acc. diff. & Train ratio & Infer. ratio \\
\midrule
MiniRocket & S=2 & 6 & 1 & 8 & \baddelta{-0.81 [-2.94, +1.45]} & 2.40 [1.37, 3.72] & \gooddelta{$\times$0.87} & \gooddelta{$\times$0.88} \\
MiniRocket & S=3 & 10 & 1 & 4 & \gooddelta{+1.11 [-0.43, +2.07]} & 1.75 [0.95, 4.44] & \gooddelta{$\times$0.67} & \gooddelta{$\times$0.57} \\
MiniRocket & S=4 & 7 & 4 & 4 & \gooddelta{+0.28 [-0.70, +2.68]} & 1.35 [0.65, 5.72] & \gooddelta{$\times$0.59} & \gooddelta{$\times$0.44} \\
\midrule
MultiRocket & S=2 & 5 & 5 & 5 & \baddelta{+0.00 [-1.64, +1.20]} & 1.33 [0.31, 2.27] & \gooddelta{$\times$0.78} & \gooddelta{$\times$0.77} \\
MultiRocket & S=3 & 5 & 4 & 6 & \baddelta{-0.27 [-2.44, +1.64]} & 1.95 [0.85, 4.69] & \gooddelta{$\times$0.49} & \gooddelta{$\times$0.51} \\
MultiRocket & S=4 & 5 & 0 & 10 & \baddelta{-0.97 [-3.23, +2.10]} & 2.67 [1.28, 4.28] & \gooddelta{$\times$0.34} & \gooddelta{$\times$0.35} \\
\midrule
CNN Classifier & S=2 & 5 & 3 & 7 & \baddelta{-0.06 [-1.94, +2.20]} & 2.50 [0.77, 3.40] & \gooddelta{$\times$0.99} & \baddelta{$\times$1.03} \\
CNN Classifier & S=3 & 7 & 1 & 7 & \gooddelta{+0.32 [-2.63, +1.21]} & 2.60 [1.04, 5.65] & \gooddelta{$\times$0.90} & \baddelta{$\times$1.13} \\
CNN Classifier & S=4 & 7 & 1 & 7 & \baddelta{-0.13 [-3.07, +2.81]} & 3.71 [1.59, 8.51] & \baddelta{$\times$1.04} & \baddelta{$\times$1.21} \\
\midrule
FCN Classifier & S=2 & 10 & 1 & 4 & \gooddelta{+1.02 [-0.41, +2.27]} & 1.33 [0.92, 2.60] & \gooddelta{$\times$0.80} & \gooddelta{$\times$0.95} \\
FCN Classifier & S=3 & 8 & 1 & 6 & \gooddelta{+0.78 [-2.31, +3.85]} & 3.05 [1.73, 4.67] & \gooddelta{$\times$0.95} & \baddelta{$\times$1.10} \\
FCN Classifier & S=4 & 5 & 2 & 8 & \baddelta{-1.44 [-3.96, +2.07]} & 3.67 [1.50, 6.11] & \gooddelta{$\times$0.95} & \baddelta{$\times$1.12} \\
\bottomrule
\end{tabular}
}
\endgroup
\end{table}
 Each cell gives the mean test accuracy across the five seeds, with the corresponding sample standard deviation in parentheses. Dataset names are followed by the time-series length in parentheses.

\FloatBarrier
\begingroup
\scriptsize
\setlength{\tabcolsep}{2.7pt}
\renewcommand{\arraystretch}{1.03}
\setlength{\LTcapwidth}{\linewidth}
\begin{longtable}{p{4.8cm}cccc}
\caption{Per-dataset UCR benchmark results for MiniRocket. Each cell reports the mean test accuracy with standard deviation in parentheses over the five seeds for the corresponding ROMAN setting. Dataset names are followed by the time-series length in parentheses.}\label{tab:complete_results_ucr_minirocket}\\
\toprule
Dataset & \multicolumn{4}{c}{MiniRocket} \\
\cmidrule(lr){2-5}
& S=1 & S=2 & S=3 & S=4 \\
\midrule
\endfirsthead
\caption[]{UCR MiniRocket benchmark results (continued)}\\
\toprule
Dataset & \multicolumn{4}{c}{MiniRocket} \\
\cmidrule(lr){2-5}
& S=1 & S=2 & S=3 & S=4 \\
\midrule
\endhead
\midrule
\multicolumn{5}{r}{Continued on next page}\\
\endfoot
\bottomrule
\endlastfoot
ACSF1 (1460) & 0.908 (0.008) & 0.896 (0.011) & 0.860 (0.010) & 0.830 (0.012) \\
AllGestureWiimoteX (385) & 0.643 (0.008) & 0.691 (0.008) & 0.707 (0.016) & 0.667 (0.011) \\
AllGestureWiimoteY (385) & 0.684 (0.004) & 0.713 (0.008) & 0.725 (0.007) & 0.687 (0.009) \\
AllGestureWiimoteZ (385) & 0.623 (0.011) & 0.626 (0.010) & 0.606 (0.007) & 0.581 (0.008) \\
AsphaltObstacles (736) & 0.856 (0.006) & 0.883 (0.006) & 0.887 (0.006) & 0.886 (0.007) \\
Beef (470) & 0.853 (0.018) & 0.820 (0.030) & 0.840 (0.028) & 0.853 (0.018) \\
BeetleFly (512) & 0.920 (0.027) & 0.950 (0.000) & 0.940 (0.022) & 0.890 (0.065) \\
BirdChicken (512) & 0.900 (0.000) & 0.920 (0.027) & 0.900 (0.079) & 0.780 (0.057) \\
Car (577) & 0.910 (0.015) & 0.897 (0.018) & 0.933 (0.000) & 0.920 (0.018) \\
CinCECGTorso (1639) & 0.808 (0.013) & 0.906 (0.017) & 0.950 (0.005) & 0.959 (0.010) \\
Coffee (286) & 1.000 (0.000) & 1.000 (0.000) & 1.000 (0.000) & 1.000 (0.000) \\
Computers (720) & 0.716 (0.012) & 0.738 (0.004) & 0.729 (0.010) & 0.716 (0.012) \\
CricketX (300) & 0.795 (0.008) & 0.769 (0.009) & 0.738 (0.013) & 0.717 (0.015) \\
CricketY (300) & 0.805 (0.007) & 0.793 (0.007) & 0.770 (0.007) & 0.759 (0.010) \\
CricketZ (300) & 0.781 (0.009) & 0.785 (0.013) & 0.756 (0.004) & 0.738 (0.004) \\
DiatomSizeReduction (345) & 0.933 (0.008) & 0.944 (0.003) & 0.976 (0.002) & 0.976 (0.003) \\
DodgerLoopDay (288) & 0.647 (0.024) & 0.645 (0.036) & 0.662 (0.023) & 0.610 (0.029) \\
DodgerLoopGame (288) & 0.868 (0.006) & 0.923 (0.006) & 0.930 (0.004) & 0.919 (0.006) \\
DodgerLoopWeekend (288) & 0.978 (0.000) & 0.986 (0.000) & 0.986 (0.000) & 0.986 (0.000) \\
EOGHorizontalSignal (1250) & 0.587 (0.006) & 0.603 (0.008) & 0.605 (0.011) & 0.588 (0.023) \\
EOGVerticalSignal (1250) & 0.533 (0.011) & 0.538 (0.015) & 0.535 (0.011) & 0.519 (0.014) \\
Earthquakes (512) & 0.748 (0.000) & 0.748 (0.000) & 0.748 (0.000) & 0.748 (0.000) \\
EthanolLevel (1751) & 0.566 (0.009) & 0.528 (0.006) & 0.558 (0.006) & 0.562 (0.010) \\
FaceFour (350) & 0.943 (0.000) & 1.000 (0.000) & 0.998 (0.005) & 0.984 (0.013) \\
FiftyWords (270) & 0.805 (0.003) & 0.812 (0.005) & 0.801 (0.005) & 0.768 (0.007) \\
Fish (463) & 0.959 (0.005) & 0.967 (0.007) & 0.959 (0.015) & 0.960 (0.009) \\
FordA (500) & 0.949 (0.003) & 0.940 (0.002) & 0.929 (0.005) & 0.913 (0.008) \\
FordB (500) & 0.825 (0.006) & 0.808 (0.013) & 0.791 (0.009) & 0.777 (0.006) \\
FreezerRegularTrain (301) & 0.999 (0.000) & 0.998 (0.000) & 0.998 (0.000) & 0.997 (0.001) \\
FreezerSmallTrain (301) & 0.940 (0.001) & 0.931 (0.004) & 0.929 (0.008) & 0.940 (0.016) \\
GestureMidAirD1 (360) & 0.718 (0.013) & 0.765 (0.010) & 0.762 (0.009) & 0.749 (0.018) \\
GestureMidAirD2 (360) & 0.614 (0.013) & 0.668 (0.006) & 0.648 (0.006) & 0.662 (0.018) \\
GestureMidAirD3 (360) & 0.443 (0.021) & 0.508 (0.022) & 0.523 (0.020) & 0.549 (0.017) \\
GesturePebbleZ1 (455) & 0.902 (0.005) & 0.895 (0.004) & 0.893 (0.003) & 0.878 (0.004) \\
GesturePebbleZ2 (455) & 0.876 (0.017) & 0.867 (0.012) & 0.843 (0.010) & 0.833 (0.010) \\
Ham (431) & 0.750 (0.025) & 0.699 (0.014) & 0.728 (0.009) & 0.741 (0.018) \\
HandOutlines (2709) & 0.936 (0.005) & 0.937 (0.004) & 0.936 (0.007) & 0.930 (0.006) \\
Haptics (1092) & 0.527 (0.012) & 0.499 (0.005) & 0.501 (0.007) & 0.487 (0.013) \\
Herring (512) & 0.600 (0.014) & 0.597 (0.007) & 0.609 (0.016) & 0.606 (0.017) \\
HouseTwenty (2000) & 0.955 (0.005) & 0.963 (0.005) & 0.971 (0.008) & 0.968 (0.004) \\
InlineSkate (1882) & 0.415 (0.005) & 0.363 (0.004) & 0.379 (0.014) & 0.371 (0.026) \\
InsectEPGRegularTrain (601) & 1.000 (0.000) & 1.000 (0.000) & 1.000 (0.000) & 1.000 (0.000) \\
InsectEPGSmallTrain (601) & 0.998 (0.002) & 1.000 (0.000) & 1.000 (0.000) & 1.000 (0.000) \\
InsectWingbeatSound (256) & 0.666 (0.001) & 0.669 (0.003) & 0.667 (0.003) & 0.670 (0.004) \\
LargeKitchenAppliances (720) & 0.822 (0.007) & 0.846 (0.012) & 0.855 (0.008) & 0.814 (0.009) \\
Lightning2 (637) & 0.744 (0.009) & 0.761 (0.019) & 0.708 (0.014) & 0.715 (0.025) \\
Lightning7 (319) & 0.770 (0.015) & 0.816 (0.008) & 0.803 (0.025) & 0.762 (0.023) \\
Mallat (1024) & 0.932 (0.011) & 0.927 (0.010) & 0.934 (0.005) & 0.922 (0.009) \\
Meat (448) & 0.973 (0.019) & 0.990 (0.009) & 0.983 (0.017) & 0.973 (0.022) \\
MixedShapesRegularTrain (1024) & 0.967 (0.002) & 0.966 (0.001) & 0.963 (0.001) & 0.961 (0.002) \\
MixedShapesSmallTrain (1024) & 0.944 (0.002) & 0.928 (0.002) & 0.925 (0.003) & 0.922 (0.004) \\
NonInvasiveFetalECGThorax1 (750) & 0.931 (0.002) & 0.929 (0.002) & 0.925 (0.004) & 0.926 (0.002) \\
NonInvasiveFetalECGThorax2 (750) & 0.953 (0.001) & 0.949 (0.003) & 0.950 (0.001) & 0.949 (0.004) \\
OSULeaf (427) & 0.921 (0.009) & 0.859 (0.015) & 0.772 (0.015) & 0.737 (0.013) \\
OliveOil (570) & 0.940 (0.043) & 0.940 (0.028) & 0.913 (0.018) & 0.953 (0.018) \\
PLAID (1344) & 0.944 (0.004) & 0.935 (0.004) & 0.939 (0.010) & 0.930 (0.004) \\
Phoneme (1024) & 0.246 (0.004) & 0.273 (0.005) & 0.275 (0.001) & 0.266 (0.004) \\
PickupGestureWiimoteZ (361) & 0.744 (0.026) & 0.800 (0.000) & 0.776 (0.022) & 0.748 (0.036) \\
PigAirwayPressure (2000) & 0.218 (0.012) & 0.099 (0.004) & 0.096 (0.003) & 0.083 (0.007) \\
PigArtPressure (2000) & 0.963 (0.005) & 0.573 (0.010) & 0.434 (0.017) & 0.346 (0.016) \\
PigCVP (2000) & 0.732 (0.072) & 0.403 (0.011) & 0.256 (0.024) & 0.227 (0.010) \\
RefrigerationDevices (720) & 0.469 (0.006) & 0.487 (0.011) & 0.481 (0.009) & 0.499 (0.013) \\
Rock (2844) & 0.780 (0.024) & 0.884 (0.038) & 0.820 (0.014) & 0.808 (0.011) \\
ScreenType (720) & 0.457 (0.011) & 0.460 (0.013) & 0.450 (0.014) & 0.458 (0.022) \\
SemgHandGenderCh2 (1500) & 0.885 (0.013) & 0.933 (0.006) & 0.953 (0.006) & 0.942 (0.004) \\
SemgHandMovementCh2 (1500) & 0.672 (0.015) & 0.724 (0.015) & 0.743 (0.005) & 0.742 (0.011) \\
SemgHandSubjectCh2 (1500) & 0.859 (0.010) & 0.891 (0.013) & 0.904 (0.008) & 0.922 (0.007) \\
ShakeGestureWiimoteZ (385) & 0.832 (0.018) & 0.868 (0.023) & 0.868 (0.011) & 0.844 (0.017) \\
ShapeletSim (500) & 0.876 (0.026) & 0.612 (0.028) & 0.688 (0.041) & 0.689 (0.026) \\
ShapesAll (512) & 0.903 (0.006) & 0.887 (0.003) & 0.874 (0.002) & 0.860 (0.006) \\
SmallKitchenAppliances (720) & 0.798 (0.003) & 0.820 (0.007) & 0.829 (0.007) & 0.813 (0.003) \\
StarLightCurves (1024) & 0.982 (0.001) & 0.981 (0.000) & 0.981 (0.000) & 0.981 (0.001) \\
Symbols (398) & 0.977 (0.001) & 0.963 (0.001) & 0.965 (0.002) & 0.958 (0.004) \\
ToeSegmentation1 (277) & 0.952 (0.008) & 0.910 (0.020) & 0.832 (0.023) & 0.710 (0.046) \\
ToeSegmentation2 (343) & 0.912 (0.014) & 0.868 (0.006) & 0.863 (0.003) & 0.832 (0.015) \\
Trace (275) & 1.000 (0.000) & 1.000 (0.000) & 1.000 (0.000) & 0.996 (0.005) \\
UWaveGestureLibraryAll (945) & 0.969 (0.002) & 0.976 (0.002) & 0.975 (0.001) & 0.975 (0.002) \\
UWaveGestureLibraryX (315) & 0.842 (0.005) & 0.845 (0.002) & 0.840 (0.001) & 0.830 (0.008) \\
UWaveGestureLibraryY (315) & 0.750 (0.002) & 0.758 (0.004) & 0.753 (0.003) & 0.740 (0.010) \\
UWaveGestureLibraryZ (315) & 0.792 (0.003) & 0.792 (0.002) & 0.780 (0.001) & 0.752 (0.005) \\
WordSynonyms (270) & 0.704 (0.001) & 0.721 (0.005) & 0.713 (0.009) & 0.686 (0.006) \\
Worms (900) & 0.719 (0.020) & 0.719 (0.012) & 0.694 (0.022) & 0.649 (0.030) \\
WormsTwoClass (900) & 0.758 (0.020) & 0.753 (0.021) & 0.696 (0.040) & 0.686 (0.027) \\
\end{longtable}
\endgroup

\begingroup
\scriptsize
\setlength{\tabcolsep}{2.7pt}
\renewcommand{\arraystretch}{1.03}
\setlength{\LTcapwidth}{\linewidth}
\begin{longtable}{p{4.8cm}cccc}
\caption{Per-dataset UCR benchmark results for MultiRocket.}\label{tab:complete_results_ucr_multirocket}\\
\toprule
Dataset & \multicolumn{4}{c}{MultiRocket} \\
\cmidrule(lr){2-5}
& S=1 & S=2 & S=3 & S=4 \\
\midrule
\endfirsthead
\caption[]{UCR MultiRocket benchmark results (continued)}\\
\toprule
Dataset & \multicolumn{4}{c}{MultiRocket} \\
\cmidrule(lr){2-5}
& S=1 & S=2 & S=3 & S=4 \\
\midrule
\endhead
\midrule
\multicolumn{5}{r}{Continued on next page}\\
\endfoot
\bottomrule
\endlastfoot
ACSF1 (1460) & 0.884 (0.011) & 0.838 (0.013) & 0.836 (0.009) & 0.788 (0.022) \\
AllGestureWiimoteX (385) & 0.719 (0.006) & 0.721 (0.006) & 0.710 (0.019) & 0.669 (0.019) \\
AllGestureWiimoteY (385) & 0.745 (0.010) & 0.757 (0.010) & 0.745 (0.017) & 0.705 (0.007) \\
AllGestureWiimoteZ (385) & 0.682 (0.015) & 0.661 (0.008) & 0.665 (0.012) & 0.623 (0.017) \\
AsphaltObstacles (736) & 0.829 (0.012) & 0.847 (0.019) & 0.862 (0.009) & 0.856 (0.019) \\
Beef (470) & 0.707 (0.028) & 0.687 (0.051) & 0.720 (0.069) & 0.793 (0.028) \\
BeetleFly (512) & 0.890 (0.042) & 0.880 (0.076) & 0.870 (0.057) & 0.890 (0.042) \\
BirdChicken (512) & 0.940 (0.022) & 0.900 (0.071) & 0.870 (0.045) & 0.860 (0.082) \\
Car (577) & 0.887 (0.007) & 0.883 (0.012) & 0.890 (0.019) & 0.897 (0.007) \\
CinCECGTorso (1639) & 0.861 (0.005) & 0.937 (0.038) & 0.973 (0.017) & 0.978 (0.021) \\
Coffee (286) & 1.000 (0.000) & 1.000 (0.000) & 1.000 (0.000) & 1.000 (0.000) \\
Computers (720) & 0.719 (0.010) & 0.726 (0.022) & 0.704 (0.005) & 0.686 (0.013) \\
CricketX (300) & 0.762 (0.011) & 0.771 (0.010) & 0.757 (0.008) & 0.718 (0.016) \\
CricketY (300) & 0.784 (0.010) & 0.782 (0.011) & 0.775 (0.015) & 0.751 (0.017) \\
CricketZ (300) & 0.817 (0.008) & 0.803 (0.013) & 0.768 (0.011) & 0.745 (0.008) \\
DiatomSizeReduction (345) & 0.956 (0.004) & 0.954 (0.019) & 0.951 (0.012) & 0.963 (0.015) \\
DodgerLoopDay (288) & 0.470 (0.029) & 0.542 (0.046) & 0.568 (0.053) & 0.597 (0.037) \\
DodgerLoopGame (288) & 0.893 (0.017) & 0.896 (0.018) & 0.888 (0.036) & 0.884 (0.007) \\
DodgerLoopWeekend (288) & 0.975 (0.004) & 0.978 (0.005) & 0.986 (0.000) & 0.986 (0.000) \\
EOGHorizontalSignal (1250) & 0.623 (0.039) & 0.635 (0.021) & 0.615 (0.013) & 0.606 (0.018) \\
EOGVerticalSignal (1250) & 0.587 (0.018) & 0.608 (0.012) & 0.599 (0.020) & 0.568 (0.032) \\
Earthquakes (512) & 0.715 (0.023) & 0.706 (0.039) & 0.741 (0.013) & 0.721 (0.027) \\
EthanolLevel (1751) & 0.580 (0.013) & 0.578 (0.017) & 0.561 (0.019) & 0.589 (0.020) \\
FaceFour (350) & 0.927 (0.006) & 0.936 (0.017) & 0.968 (0.015) & 0.975 (0.012) \\
FiftyWords (270) & 0.829 (0.008) & 0.827 (0.008) & 0.795 (0.012) & 0.767 (0.005) \\
Fish (463) & 0.981 (0.007) & 0.953 (0.009) & 0.960 (0.011) & 0.947 (0.011) \\
FordA (500) & 0.900 (0.009) & 0.890 (0.008) & 0.898 (0.002) & 0.877 (0.012) \\
FordB (500) & 0.757 (0.014) & 0.760 (0.008) & 0.758 (0.017) & 0.751 (0.015) \\
FreezerRegularTrain (301) & 0.998 (0.000) & 0.995 (0.001) & 0.996 (0.001) & 0.994 (0.002) \\
FreezerSmallTrain (301) & 0.959 (0.004) & 0.930 (0.015) & 0.846 (0.025) & 0.842 (0.051) \\
GestureMidAirD1 (360) & 0.705 (0.042) & 0.714 (0.013) & 0.755 (0.010) & 0.746 (0.020) \\
GestureMidAirD2 (360) & 0.602 (0.020) & 0.652 (0.015) & 0.626 (0.024) & 0.625 (0.027) \\
GestureMidAirD3 (360) & 0.406 (0.032) & 0.438 (0.031) & 0.463 (0.024) & 0.491 (0.010) \\
GesturePebbleZ1 (455) & 0.924 (0.006) & 0.913 (0.000) & 0.893 (0.003) & 0.884 (0.004) \\
GesturePebbleZ2 (455) & 0.854 (0.004) & 0.839 (0.006) & 0.792 (0.019) & 0.827 (0.015) \\
Ham (431) & 0.750 (0.008) & 0.741 (0.017) & 0.733 (0.021) & 0.709 (0.020) \\
HandOutlines (2709) & 0.894 (0.010) & 0.903 (0.013) & 0.912 (0.017) & 0.916 (0.012) \\
Haptics (1092) & 0.519 (0.017) & 0.506 (0.016) & 0.518 (0.015) & 0.513 (0.025) \\
Herring (512) & 0.647 (0.036) & 0.650 (0.034) & 0.619 (0.070) & 0.637 (0.026) \\
HouseTwenty (2000) & 0.946 (0.020) & 0.926 (0.016) & 0.931 (0.021) & 0.938 (0.033) \\
InlineSkate (1882) & 0.470 (0.009) & 0.444 (0.024) & 0.414 (0.009) & 0.414 (0.007) \\
InsectEPGRegularTrain (601) & 1.000 (0.000) & 1.000 (0.000) & 1.000 (0.000) & 1.000 (0.000) \\
InsectEPGSmallTrain (601) & 1.000 (0.000) & 1.000 (0.000) & 1.000 (0.000) & 1.000 (0.000) \\
InsectWingbeatSound (256) & 0.640 (0.003) & 0.646 (0.006) & 0.652 (0.006) & 0.643 (0.007) \\
LargeKitchenAppliances (720) & 0.765 (0.023) & 0.803 (0.015) & 0.818 (0.011) & 0.776 (0.023) \\
Lightning2 (637) & 0.734 (0.014) & 0.725 (0.050) & 0.734 (0.036) & 0.725 (0.024) \\
Lightning7 (319) & 0.775 (0.016) & 0.737 (0.047) & 0.756 (0.042) & 0.726 (0.036) \\
Mallat (1024) & 0.952 (0.003) & 0.953 (0.013) & 0.960 (0.005) & 0.950 (0.023) \\
Meat (448) & 0.957 (0.009) & 0.947 (0.018) & 0.963 (0.032) & 0.957 (0.025) \\
MixedShapesRegularTrain (1024) & 0.974 (0.002) & 0.969 (0.005) & 0.968 (0.001) & 0.967 (0.002) \\
MixedShapesSmallTrain (1024) & 0.948 (0.002) & 0.944 (0.003) & 0.933 (0.004) & 0.932 (0.003) \\
NonInvasiveFetalECGThorax1 (750) & 0.933 (0.004) & 0.918 (0.009) & 0.920 (0.006) & 0.925 (0.006) \\
NonInvasiveFetalECGThorax2 (750) & 0.943 (0.004) & 0.937 (0.011) & 0.944 (0.006) & 0.942 (0.004) \\
OSULeaf (427) & 0.948 (0.010) & 0.869 (0.015) & 0.793 (0.015) & 0.765 (0.013) \\
OliveOil (570) & 0.880 (0.051) & 0.873 (0.064) & 0.893 (0.043) & 0.913 (0.038) \\
PLAID (1344) & 0.947 (0.007) & 0.945 (0.004) & 0.943 (0.008) & 0.947 (0.004) \\
Phoneme (1024) & 0.324 (0.006) & 0.333 (0.010) & 0.323 (0.004) & 0.319 (0.003) \\
PickupGestureWiimoteZ (361) & 0.788 (0.023) & 0.756 (0.036) & 0.756 (0.036) & 0.764 (0.048) \\
PigAirwayPressure (2000) & 0.778 (0.039) & 0.290 (0.018) & 0.185 (0.011) & 0.157 (0.008) \\
PigArtPressure (2000) & 0.969 (0.003) & 0.887 (0.031) & 0.776 (0.050) & 0.687 (0.027) \\
PigCVP (2000) & 0.855 (0.005) & 0.736 (0.023) & 0.613 (0.026) & 0.499 (0.011) \\
RefrigerationDevices (720) & 0.517 (0.026) & 0.510 (0.019) & 0.515 (0.019) & 0.525 (0.023) \\
Rock (2844) & 0.792 (0.027) & 0.856 (0.022) & 0.864 (0.030) & 0.868 (0.027) \\
ScreenType (720) & 0.462 (0.012) & 0.439 (0.017) & 0.449 (0.016) & 0.442 (0.029) \\
SemgHandGenderCh2 (1500) & 0.930 (0.004) & 0.936 (0.008) & 0.932 (0.008) & 0.938 (0.006) \\
SemgHandMovementCh2 (1500) & 0.737 (0.012) & 0.749 (0.025) & 0.746 (0.024) & 0.760 (0.014) \\
SemgHandSubjectCh2 (1500) & 0.858 (0.014) & 0.881 (0.011) & 0.883 (0.008) & 0.906 (0.013) \\
ShakeGestureWiimoteZ (385) & 0.912 (0.018) & 0.888 (0.011) & 0.896 (0.017) & 0.876 (0.054) \\
ShapeletSim (500) & 1.000 (0.000) & 0.842 (0.046) & 0.718 (0.093) & 0.651 (0.073) \\
ShapesAll (512) & 0.921 (0.003) & 0.889 (0.005) & 0.871 (0.006) & 0.866 (0.008) \\
SmallKitchenAppliances (720) & 0.765 (0.015) & 0.779 (0.013) & 0.780 (0.007) & 0.779 (0.011) \\
StarLightCurves (1024) & 0.980 (0.001) & 0.980 (0.001) & 0.981 (0.000) & 0.981 (0.001) \\
Symbols (398) & 0.983 (0.001) & 0.978 (0.003) & 0.974 (0.005) & 0.966 (0.007) \\
ToeSegmentation1 (277) & 0.919 (0.006) & 0.901 (0.033) & 0.867 (0.038) & 0.796 (0.021) \\
ToeSegmentation2 (343) & 0.880 (0.009) & 0.888 (0.028) & 0.848 (0.023) & 0.858 (0.015) \\
Trace (275) & 1.000 (0.000) & 1.000 (0.000) & 1.000 (0.000) & 0.994 (0.009) \\
UWaveGestureLibraryAll (945) & 0.975 (0.001) & 0.974 (0.001) & 0.973 (0.001) & 0.972 (0.001) \\
UWaveGestureLibraryX (315) & 0.852 (0.004) & 0.847 (0.003) & 0.837 (0.005) & 0.832 (0.003) \\
UWaveGestureLibraryY (315) & 0.781 (0.001) & 0.767 (0.003) & 0.765 (0.006) & 0.746 (0.006) \\
UWaveGestureLibraryZ (315) & 0.795 (0.002) & 0.786 (0.008) & 0.776 (0.005) & 0.757 (0.003) \\
WordSynonyms (270) & 0.737 (0.005) & 0.743 (0.009) & 0.708 (0.010) & 0.686 (0.010) \\
Worms (900) & 0.730 (0.027) & 0.709 (0.020) & 0.714 (0.018) & 0.717 (0.023) \\
WormsTwoClass (900) & 0.766 (0.033) & 0.756 (0.025) & 0.771 (0.028) & 0.766 (0.021) \\
\end{longtable}
\endgroup

\begingroup
\scriptsize
\setlength{\tabcolsep}{2.7pt}
\renewcommand{\arraystretch}{1.03}
\setlength{\LTcapwidth}{\linewidth}
\begin{longtable}{p{4.8cm}cccc}
\caption{Per-dataset UCR benchmark results for CNN.}\label{tab:complete_results_ucr_cnn}\\
\toprule
Dataset & \multicolumn{4}{c}{CNN} \\
\cmidrule(lr){2-5}
& S=1 & S=2 & S=3 & S=4 \\
\midrule
\endfirsthead
\caption[]{UCR CNN benchmark results (continued)}\\
\toprule
Dataset & \multicolumn{4}{c}{CNN} \\
\cmidrule(lr){2-5}
& S=1 & S=2 & S=3 & S=4 \\
\midrule
\endhead
\midrule
\multicolumn{5}{r}{Continued on next page}\\
\endfoot
\bottomrule
\endlastfoot
ACSF1 (1460) & 0.228 (0.057) & 0.176 (0.058) & 0.200 (0.067) & 0.316 (0.141) \\
AllGestureWiimoteX (385) & 0.342 (0.053) & 0.426 (0.037) & 0.424 (0.039) & 0.388 (0.034) \\
AllGestureWiimoteY (385) & 0.286 (0.112) & 0.392 (0.059) & 0.436 (0.052) & 0.406 (0.051) \\
AllGestureWiimoteZ (385) & 0.251 (0.121) & 0.266 (0.059) & 0.307 (0.067) & 0.339 (0.037) \\
AsphaltObstacles (736) & 0.756 (0.024) & 0.765 (0.022) & 0.760 (0.032) & 0.753 (0.043) \\
Beef (470) & 0.700 (0.041) & 0.720 (0.030) & 0.660 (0.098) & 0.687 (0.077) \\
BeetleFly (512) & 0.700 (0.184) & 0.690 (0.182) & 0.700 (0.137) & 0.700 (0.100) \\
BirdChicken (512) & 0.630 (0.045) & 0.600 (0.100) & 0.610 (0.055) & 0.630 (0.057) \\
Car (577) & 0.687 (0.007) & 0.683 (0.187) & 0.760 (0.030) & 0.750 (0.000) \\
CinCECGTorso (1639) & 0.546 (0.184) & 0.503 (0.229) & 0.518 (0.182) & 0.430 (0.175) \\
Coffee (286) & 0.979 (0.020) & 0.993 (0.016) & 0.971 (0.016) & 0.971 (0.030) \\
Computers (720) & 0.510 (0.044) & 0.508 (0.035) & 0.503 (0.039) & 0.502 (0.027) \\
CricketX (300) & 0.502 (0.026) & 0.507 (0.028) & 0.476 (0.022) & 0.441 (0.035) \\
CricketY (300) & 0.481 (0.022) & 0.544 (0.027) & 0.487 (0.016) & 0.439 (0.035) \\
CricketZ (300) & 0.485 (0.056) & 0.512 (0.026) & 0.480 (0.025) & 0.439 (0.022) \\
DiatomSizeReduction (345) & -- & -- & -- & -- \\
DodgerLoopDay (288) & 0.595 (0.033) & 0.603 (0.037) & 0.578 (0.116) & 0.620 (0.031) \\
DodgerLoopGame (288) & 0.619 (0.187) & 0.710 (0.125) & 0.767 (0.144) & 0.787 (0.112) \\
DodgerLoopWeekend (288) & 0.984 (0.003) & 0.980 (0.006) & 0.972 (0.006) & 0.974 (0.004) \\
EOGHorizontalSignal (1250) & 0.409 (0.031) & 0.414 (0.020) & 0.425 (0.027) & 0.429 (0.017) \\
EOGVerticalSignal (1250) & 0.360 (0.037) & 0.393 (0.020) & 0.378 (0.055) & 0.374 (0.017) \\
Earthquakes (512) & 0.750 (0.003) & 0.748 (0.000) & 0.748 (0.000) & 0.748 (0.000) \\
EthanolLevel (1751) & 0.258 (0.013) & 0.248 (0.014) & 0.258 (0.008) & 0.251 (0.002) \\
FaceFour (350) & -- & -- & -- & -- \\
FiftyWords (270) & -- & -- & -- & -- \\
Fish (463) & 0.539 (0.035) & 0.715 (0.064) & 0.723 (0.055) & 0.798 (0.022) \\
FordA (500) & 0.894 (0.024) & 0.893 (0.011) & 0.852 (0.015) & 0.732 (0.017) \\
FordB (500) & 0.703 (0.019) & 0.686 (0.027) & 0.693 (0.004) & 0.680 (0.018) \\
FreezerRegularTrain (301) & 0.932 (0.097) & 0.927 (0.091) & 0.954 (0.042) & 0.949 (0.010) \\
FreezerSmallTrain (301) & 0.745 (0.030) & 0.733 (0.021) & 0.771 (0.021) & 0.733 (0.027) \\
GestureMidAirD1 (360) & 0.474 (0.043) & 0.472 (0.053) & 0.525 (0.040) & 0.511 (0.020) \\
GestureMidAirD2 (360) & 0.489 (0.046) & 0.457 (0.036) & 0.488 (0.052) & 0.506 (0.050) \\
GestureMidAirD3 (360) & 0.257 (0.071) & 0.297 (0.031) & 0.268 (0.045) & 0.231 (0.036) \\
GesturePebbleZ1 (455) & 0.776 (0.042) & 0.834 (0.028) & 0.842 (0.027) & 0.824 (0.014) \\
GesturePebbleZ2 (455) & 0.667 (0.052) & 0.715 (0.037) & 0.790 (0.064) & 0.733 (0.041) \\
Ham (431) & 0.705 (0.013) & 0.674 (0.049) & 0.661 (0.037) & 0.714 (0.007) \\
HandOutlines (2709) & 0.892 (0.014) & 0.907 (0.006) & 0.898 (0.009) & 0.906 (0.006) \\
Haptics (1092) & 0.388 (0.021) & 0.401 (0.033) & 0.390 (0.017) & 0.402 (0.026) \\
Herring (512) & 0.594 (0.000) & 0.609 (0.031) & 0.591 (0.026) & 0.597 (0.007) \\
HouseTwenty (2000) & 0.666 (0.140) & 0.689 (0.056) & 0.676 (0.064) & 0.721 (0.080) \\
InlineSkate (1882) & 0.189 (0.026) & 0.206 (0.031) & 0.214 (0.029) & 0.188 (0.006) \\
InsectEPGRegularTrain (601) & 1.000 (0.000) & 1.000 (0.000) & 1.000 (0.000) & 1.000 (0.000) \\
InsectEPGSmallTrain (601) & 1.000 (0.000) & 1.000 (0.000) & 0.966 (0.075) & 1.000 (0.000) \\
InsectWingbeatSound (256) & 0.578 (0.017) & 0.583 (0.013) & 0.595 (0.010) & 0.593 (0.020) \\
LargeKitchenAppliances (720) & 0.515 (0.017) & 0.596 (0.046) & 0.575 (0.037) & 0.529 (0.023) \\
Lightning2 (637) & 0.643 (0.094) & 0.669 (0.117) & 0.623 (0.102) & 0.695 (0.089) \\
Lightning7 (319) & 0.633 (0.062) & 0.633 (0.111) & 0.649 (0.028) & 0.633 (0.043) \\
Mallat (1024) & -- & -- & -- & -- \\
Meat (448) & 0.873 (0.032) & 0.903 (0.040) & 0.817 (0.031) & 0.900 (0.029) \\
MixedShapesRegularTrain (1024) & 0.813 (0.043) & 0.827 (0.043) & 0.839 (0.033) & 0.838 (0.012) \\
MixedShapesSmallTrain (1024) & 0.724 (0.030) & 0.740 (0.063) & 0.728 (0.075) & 0.742 (0.027) \\
NonInvasiveFetalECGThorax1 (750) & 0.875 (0.004) & 0.806 (0.027) & 0.850 (0.010) & 0.847 (0.015) \\
NonInvasiveFetalECGThorax2 (750) & 0.915 (0.009) & 0.870 (0.012) & 0.893 (0.008) & 0.879 (0.010) \\
OSULeaf (427) & 0.379 (0.081) & 0.450 (0.006) & 0.411 (0.054) & 0.412 (0.026) \\
OliveOil (570) & 0.380 (0.045) & 0.400 (0.000) & 0.380 (0.045) & 0.380 (0.045) \\
PLAID (1344) & 0.342 (0.060) & 0.358 (0.075) & 0.371 (0.065) & 0.401 (0.062) \\
Phoneme (1024) & -- & -- & -- & -- \\
PickupGestureWiimoteZ (361) & 0.560 (0.040) & 0.628 (0.030) & 0.592 (0.058) & 0.640 (0.032) \\
PigAirwayPressure (2000) & -- & -- & -- & -- \\
PigArtPressure (2000) & -- & -- & -- & -- \\
PigCVP (2000) & -- & -- & -- & -- \\
RefrigerationDevices (720) & 0.354 (0.031) & 0.429 (0.057) & 0.327 (0.032) & 0.326 (0.019) \\
Rock (2844) & 0.552 (0.186) & 0.444 (0.151) & 0.540 (0.113) & 0.488 (0.171) \\
ScreenType (720) & 0.403 (0.028) & 0.419 (0.018) & 0.424 (0.043) & 0.418 (0.043) \\
SemgHandGenderCh2 (1500) & 0.901 (0.011) & 0.906 (0.013) & 0.908 (0.025) & 0.908 (0.015) \\
SemgHandMovementCh2 (1500) & 0.571 (0.024) & 0.672 (0.075) & 0.694 (0.040) & 0.731 (0.032) \\
SemgHandSubjectCh2 (1500) & 0.877 (0.017) & 0.848 (0.011) & 0.876 (0.012) & 0.888 (0.018) \\
ShakeGestureWiimoteZ (385) & 0.504 (0.082) & 0.548 (0.072) & 0.524 (0.142) & 0.596 (0.030) \\
ShapeletSim (500) & 0.487 (0.033) & 0.491 (0.017) & 0.512 (0.025) & 0.504 (0.009) \\
ShapesAll (512) & 0.519 (0.032) & 0.593 (0.021) & 0.630 (0.023) & 0.577 (0.022) \\
SmallKitchenAppliances (720) & 0.560 (0.055) & 0.641 (0.027) & 0.636 (0.028) & 0.640 (0.019) \\
StarLightCurves (1024) & 0.908 (0.028) & 0.935 (0.013) & 0.932 (0.005) & 0.924 (0.010) \\
Symbols (398) & -- & -- & -- & -- \\
ToeSegmentation1 (277) & 0.518 (0.029) & 0.508 (0.030) & 0.535 (0.033) & 0.515 (0.033) \\
ToeSegmentation2 (343) & 0.514 (0.230) & 0.331 (0.210) & 0.468 (0.264) & 0.409 (0.247) \\
Trace (275) & 0.944 (0.011) & 0.772 (0.149) & 0.754 (0.122) & 0.626 (0.097) \\
UWaveGestureLibraryAll (945) & 0.909 (0.008) & 0.910 (0.013) & 0.918 (0.007) & 0.907 (0.011) \\
UWaveGestureLibraryX (315) & 0.691 (0.024) & 0.727 (0.010) & 0.718 (0.013) & 0.695 (0.009) \\
UWaveGestureLibraryY (315) & 0.619 (0.012) & 0.629 (0.007) & 0.639 (0.010) & 0.633 (0.011) \\
UWaveGestureLibraryZ (315) & 0.638 (0.020) & 0.654 (0.028) & 0.650 (0.012) & 0.642 (0.011) \\
WordSynonyms (270) & -- & -- & -- & -- \\
Worms (900) & 0.421 (0.012) & 0.353 (0.038) & 0.366 (0.055) & 0.387 (0.030) \\
WormsTwoClass (900) & 0.558 (0.048) & 0.564 (0.042) & 0.584 (0.036) & 0.571 (0.033) \\
\end{longtable}
\endgroup

\begingroup
\scriptsize
\setlength{\tabcolsep}{2.7pt}
\renewcommand{\arraystretch}{1.03}
\setlength{\LTcapwidth}{\linewidth}
\begin{longtable}{p{4.8cm}cccc}
\caption{Per-dataset UCR benchmark results for FCN.}\label{tab:complete_results_ucr_fcn}\\
\toprule
Dataset & \multicolumn{4}{c}{FCN} \\
\cmidrule(lr){2-5}
& S=1 & S=2 & S=3 & S=4 \\
\midrule
\endfirsthead
\caption[]{UCR FCN benchmark results (continued)}\\
\toprule
Dataset & \multicolumn{4}{c}{FCN} \\
\cmidrule(lr){2-5}
& S=1 & S=2 & S=3 & S=4 \\
\midrule
\endhead
\midrule
\multicolumn{5}{r}{Continued on next page}\\
\endfoot
\bottomrule
\endlastfoot
ACSF1 (1460) & 0.660 (0.062) & 0.660 (0.060) & 0.612 (0.043) & 0.562 (0.060) \\
AllGestureWiimoteX (385) & 0.649 (0.025) & 0.638 (0.065) & 0.655 (0.033) & 0.561 (0.036) \\
AllGestureWiimoteY (385) & 0.637 (0.018) & 0.674 (0.026) & 0.659 (0.046) & 0.596 (0.059) \\
AllGestureWiimoteZ (385) & 0.644 (0.021) & 0.591 (0.022) & 0.539 (0.051) & 0.443 (0.059) \\
AsphaltObstacles (736) & 0.867 (0.019) & 0.854 (0.011) & 0.840 (0.010) & 0.828 (0.020) \\
Beef (470) & 0.340 (0.123) & 0.647 (0.112) & 0.767 (0.024) & 0.713 (0.093) \\
BeetleFly (512) & 0.750 (0.087) & 0.630 (0.104) & 0.790 (0.065) & 0.690 (0.147) \\
BirdChicken (512) & 0.660 (0.230) & 0.630 (0.179) & 0.600 (0.141) & 0.540 (0.042) \\
Car (577) & 0.447 (0.194) & 0.677 (0.140) & 0.750 (0.035) & 0.800 (0.065) \\
CinCECGTorso (1639) & 0.401 (0.056) & 0.742 (0.025) & 0.853 (0.028) & 0.753 (0.203) \\
Coffee (286) & 0.836 (0.193) & 0.807 (0.248) & 0.871 (0.230) & 0.921 (0.064) \\
Computers (720) & 0.762 (0.023) & 0.584 (0.034) & 0.514 (0.022) & 0.553 (0.051) \\
CricketX (300) & 0.724 (0.016) & 0.716 (0.024) & 0.642 (0.037) & 0.597 (0.011) \\
CricketY (300) & 0.716 (0.028) & 0.678 (0.018) & 0.676 (0.023) & 0.627 (0.015) \\
CricketZ (300) & 0.737 (0.020) & 0.693 (0.041) & 0.675 (0.018) & 0.665 (0.029) \\
DiatomSizeReduction (345) & -- & -- & -- & -- \\
DodgerLoopDay (288) & 0.380 (0.150) & 0.662 (0.025) & 0.680 (0.029) & 0.630 (0.058) \\
DodgerLoopGame (288) & 0.532 (0.085) & 0.514 (0.018) & 0.554 (0.077) & 0.701 (0.164) \\
DodgerLoopWeekend (288) & 0.942 (0.010) & 0.877 (0.122) & 0.968 (0.006) & 0.932 (0.086) \\
EOGHorizontalSignal (1250) & 0.569 (0.044) & 0.548 (0.023) & 0.537 (0.023) & 0.502 (0.031) \\
EOGVerticalSignal (1250) & 0.423 (0.024) & 0.365 (0.018) & 0.533 (0.024) & 0.521 (0.011) \\
Earthquakes (512) & 0.748 (0.000) & 0.725 (0.021) & 0.740 (0.003) & 0.706 (0.045) \\
EthanolLevel (1751) & 0.287 (0.048) & 0.618 (0.040) & 0.637 (0.056) & 0.664 (0.035) \\
FaceFour (350) & -- & -- & -- & -- \\
FiftyWords (270) & -- & -- & -- & -- \\
Fish (463) & 0.897 (0.031) & 0.859 (0.040) & 0.899 (0.020) & 0.885 (0.042) \\
FordA (500) & 0.919 (0.004) & 0.937 (0.007) & 0.932 (0.003) & 0.885 (0.009) \\
FordB (500) & 0.778 (0.010) & 0.791 (0.022) & 0.782 (0.011) & 0.721 (0.015) \\
FreezerRegularTrain (301) & 0.898 (0.128) & 0.970 (0.030) & 0.988 (0.009) & 0.986 (0.005) \\
FreezerSmallTrain (301) & 0.644 (0.120) & 0.676 (0.149) & 0.635 (0.145) & 0.709 (0.032) \\
GestureMidAirD1 (360) & 0.626 (0.043) & 0.577 (0.030) & 0.589 (0.018) & 0.569 (0.032) \\
GestureMidAirD2 (360) & 0.595 (0.025) & 0.531 (0.044) & 0.509 (0.037) & 0.542 (0.036) \\
GestureMidAirD3 (360) & 0.278 (0.043) & 0.332 (0.015) & 0.314 (0.040) & 0.322 (0.024) \\
GesturePebbleZ1 (455) & 0.817 (0.054) & 0.843 (0.042) & 0.862 (0.012) & 0.856 (0.019) \\
GesturePebbleZ2 (455) & 0.728 (0.037) & 0.851 (0.043) & 0.870 (0.053) & 0.861 (0.060) \\
Ham (431) & 0.703 (0.021) & 0.680 (0.071) & 0.676 (0.019) & 0.676 (0.023) \\
HandOutlines (2709) & 0.718 (0.032) & 0.911 (0.012) & 0.899 (0.020) & 0.911 (0.013) \\
Haptics (1092) & 0.356 (0.060) & 0.389 (0.022) & 0.403 (0.028) & 0.412 (0.034) \\
Herring (512) & 0.566 (0.020) & 0.566 (0.046) & 0.619 (0.063) & 0.581 (0.065) \\
HouseTwenty (2000) & 0.859 (0.113) & 0.914 (0.057) & 0.706 (0.182) & 0.753 (0.028) \\
InlineSkate (1882) & 0.200 (0.026) & 0.340 (0.079) & 0.352 (0.018) & 0.339 (0.009) \\
InsectEPGRegularTrain (601) & 1.000 (0.000) & 1.000 (0.000) & 1.000 (0.000) & 1.000 (0.000) \\
InsectEPGSmallTrain (601) & 1.000 (0.000) & 0.933 (0.092) & 0.722 (0.237) & 1.000 (0.000) \\
InsectWingbeatSound (256) & 0.345 (0.027) & 0.565 (0.024) & 0.575 (0.021) & 0.561 (0.021) \\
LargeKitchenAppliances (720) & 0.878 (0.012) & 0.837 (0.035) & 0.715 (0.053) & 0.594 (0.041) \\
Lightning2 (637) & 0.711 (0.099) & 0.734 (0.029) & 0.757 (0.044) & 0.682 (0.034) \\
Lightning7 (319) & 0.742 (0.023) & 0.732 (0.059) & 0.748 (0.034) & 0.652 (0.050) \\
Mallat (1024) & -- & -- & -- & -- \\
Meat (448) & 0.497 (0.123) & 0.513 (0.221) & 0.467 (0.189) & 0.623 (0.169) \\
MixedShapesRegularTrain (1024) & 0.918 (0.017) & 0.894 (0.011) & 0.907 (0.007) & 0.912 (0.003) \\
MixedShapesSmallTrain (1024) & 0.692 (0.101) & 0.830 (0.008) & 0.849 (0.030) & 0.835 (0.017) \\
NonInvasiveFetalECGThorax1 (750) & 0.887 (0.032) & 0.890 (0.011) & 0.890 (0.016) & 0.885 (0.036) \\
NonInvasiveFetalECGThorax2 (750) & 0.909 (0.008) & 0.913 (0.017) & 0.915 (0.007) & 0.921 (0.010) \\
OSULeaf (427) & 0.851 (0.036) & 0.593 (0.021) & 0.547 (0.036) & 0.521 (0.027) \\
OliveOil (570) & 0.360 (0.055) & 0.360 (0.055) & 0.587 (0.185) & 0.400 (0.000) \\
PLAID (1344) & 0.370 (0.046) & 0.331 (0.041) & 0.376 (0.037) & 0.360 (0.045) \\
Phoneme (1024) & -- & -- & -- & -- \\
PickupGestureWiimoteZ (361) & 0.624 (0.084) & 0.700 (0.032) & 0.728 (0.072) & 0.680 (0.091) \\
PigAirwayPressure (2000) & -- & -- & -- & -- \\
PigArtPressure (2000) & -- & -- & -- & -- \\
PigCVP (2000) & -- & -- & -- & -- \\
RefrigerationDevices (720) & 0.519 (0.023) & 0.498 (0.048) & 0.428 (0.048) & 0.364 (0.025) \\
Rock (2844) & 0.240 (0.040) & 0.372 (0.120) & 0.348 (0.088) & 0.480 (0.174) \\
ScreenType (720) & 0.537 (0.079) & 0.412 (0.026) & 0.430 (0.028) & 0.393 (0.021) \\
SemgHandGenderCh2 (1500) & 0.845 (0.006) & 0.948 (0.006) & 0.953 (0.010) & 0.955 (0.012) \\
SemgHandMovementCh2 (1500) & 0.498 (0.039) & 0.800 (0.025) & 0.832 (0.022) & 0.828 (0.021) \\
SemgHandSubjectCh2 (1500) & 0.604 (0.016) & 0.919 (0.014) & 0.923 (0.015) & 0.940 (0.009) \\
ShakeGestureWiimoteZ (385) & 0.856 (0.096) & 0.872 (0.030) & 0.816 (0.048) & 0.760 (0.051) \\
ShapeletSim (500) & 0.913 (0.080) & 0.624 (0.170) & 0.508 (0.017) & 0.514 (0.032) \\
ShapesAll (512) & 0.847 (0.020) & 0.762 (0.020) & 0.761 (0.016) & 0.737 (0.035) \\
SmallKitchenAppliances (720) & 0.742 (0.014) & 0.745 (0.007) & 0.669 (0.047) & 0.623 (0.044) \\
StarLightCurves (1024) & 0.966 (0.006) & 0.949 (0.007) & 0.954 (0.010) & 0.956 (0.007) \\
Symbols (398) & -- & -- & -- & -- \\
ToeSegmentation1 (277) & 0.889 (0.102) & 0.690 (0.061) & 0.646 (0.096) & 0.686 (0.055) \\
ToeSegmentation2 (343) & 0.560 (0.346) & 0.780 (0.040) & 0.738 (0.124) & 0.774 (0.042) \\
Trace (275) & 1.000 (0.000) & 1.000 (0.000) & 0.986 (0.011) & 0.934 (0.013) \\
UWaveGestureLibraryAll (945) & 0.785 (0.020) & 0.936 (0.008) & 0.958 (0.005) & 0.954 (0.008) \\
UWaveGestureLibraryX (315) & 0.743 (0.010) & 0.777 (0.011) & 0.777 (0.016) & 0.747 (0.020) \\
UWaveGestureLibraryY (315) & 0.627 (0.013) & 0.678 (0.007) & 0.680 (0.012) & 0.657 (0.015) \\
UWaveGestureLibraryZ (315) & 0.696 (0.012) & 0.697 (0.015) & 0.708 (0.018) & 0.693 (0.023) \\
WordSynonyms (270) & -- & -- & -- & -- \\
Worms (900) & 0.634 (0.039) & 0.584 (0.039) & 0.522 (0.034) & 0.504 (0.051) \\
WormsTwoClass (900) & 0.704 (0.032) & 0.595 (0.021) & 0.621 (0.035) & 0.613 (0.039) \\
\end{longtable}
\endgroup

\begingroup
\scriptsize
\setlength{\tabcolsep}{2.7pt}
\renewcommand{\arraystretch}{1.03}
\setlength{\LTcapwidth}{\linewidth}

\begin{longtable}{p{4.8cm}cccc}
\caption{Complete per-dataset UEA benchmark results for MiniRocket. Each cell reports the mean test accuracy with standard deviation in parentheses over the five seeds for the corresponding ROMAN setting. Dataset names are followed by the time-series length in parentheses.}\label{tab:complete_results_uea_minirocket}\\
\toprule
Dataset & S=1 & S=2 & S=3 & S=4 \\
\midrule
\endfirsthead
\caption[]{UEA complete benchmark results for MiniRocket (continued)}\\
\toprule
Dataset & S=1 & S=2 & S=3 & S=4 \\
\midrule
\endhead
\midrule
\multicolumn{5}{r}{Continued on next page}\\
\endfoot
\bottomrule
\endlastfoot
AsphObstCoord (736) & 0.828 (0.008) & 0.840 (0.003) & 0.844 (0.008) & 0.824 (0.005) \\
AsphPaveTypeCoord (2371) & 0.924 (0.007) & 0.935 (0.005) & 0.941 (0.004) & 0.932 (0.005) \\
AsphaltRegularityCoordinates (4201) & 0.992 (0.002) & 0.995 (0.001) & 0.998 (0.001) & 0.994 (0.001) \\
AtrialFibrillation (640) & 0.093 (0.060) & 0.053 (0.056) & 0.053 (0.056) & 0.133 (0.047) \\
Cricket (1197) & 0.961 (0.006) & 0.978 (0.008) & 0.972 (0.010) & 0.964 (0.008) \\
DuckDuckGeese (270) & 0.644 (0.036) & 0.568 (0.023) & 0.500 (0.024) & 0.464 (0.033) \\
EigenWorms (17984) & 0.925 (0.019) & 0.899 (0.026) & 0.876 (0.014) & 0.864 (0.024) \\
EthanolConcentration (1751) & 0.304 (0.011) & 0.289 (0.015) & 0.322 (0.025) & 0.373 (0.009) \\
HandMovementDirection (400) & 0.468 (0.028) & 0.543 (0.058) & 0.565 (0.040) & 0.481 (0.042) \\
Heartbeat (405) & 0.745 (0.013) & 0.730 (0.006) & 0.753 (0.018) & 0.756 (0.008) \\
MotorImagery (3000) & 0.548 (0.042) & 0.572 (0.027) & 0.572 (0.019) & 0.530 (0.032) \\
SelfRegulationSCP1 (896) & 0.911 (0.014) & 0.878 (0.031) & 0.905 (0.007) & 0.911 (0.005) \\
SelfRegulationSCP2 (1152) & 0.514 (0.025) & 0.549 (0.019) & 0.563 (0.022) & 0.579 (0.014) \\
StandWalkJump (2500) & 0.307 (0.037) & 0.267 (0.000) & 0.333 (0.115) & 0.360 (0.037) \\
UWaveGestureLibrary (315) & 0.934 (0.009) & 0.926 (0.004) & 0.931 (0.012) & 0.924 (0.011) \\
\end{longtable}

\begin{longtable}{p{4.8cm}cccc}
\caption{Complete per-dataset UEA benchmark results for MultiRocket.}\label{tab:complete_results_uea_multirocket}\\
\toprule
Dataset & S=1 & S=2 & S=3 & S=4 \\
\midrule
\endfirsthead
\caption[]{UEA complete benchmark results for MultiRocket (continued)}\\
\toprule
Dataset & S=1 & S=2 & S=3 & S=4 \\
\midrule
\endhead
\midrule
\multicolumn{5}{r}{Continued on next page}\\
\endfoot
\bottomrule
\endlastfoot
AsphObstCoord (736) & 0.773 (0.015) & 0.794 (0.014) & 0.823 (0.013) & 0.828 (0.016) \\
AsphPaveTypeCoord (2371) & 0.895 (0.010) & 0.875 (0.015) & 0.878 (0.016) & 0.885 (0.015) \\
AsphaltRegularityCoordinates (4201) & 0.989 (0.002) & 0.989 (0.003) & 0.986 (0.004) & 0.983 (0.005) \\
AtrialFibrillation (640) & 0.187 (0.056) & 0.173 (0.060) & 0.120 (0.056) & 0.147 (0.056) \\
Cricket (1197) & 0.986 (0.010) & 0.983 (0.006) & 0.983 (0.006) & 0.969 (0.012) \\
DuckDuckGeese (270) & 0.500 (0.020) & 0.436 (0.017) & 0.432 (0.054) & 0.472 (0.054) \\
EigenWorms (17984) & 0.889 (0.014) & 0.864 (0.022) & 0.873 (0.020) & 0.852 (0.013) \\
EthanolConcentration (1751) & 0.478 (0.039) & 0.392 (0.039) & 0.434 (0.016) & 0.415 (0.026) \\
HandMovementDirection (400) & 0.341 (0.050) & 0.419 (0.042) & 0.430 (0.020) & 0.438 (0.012) \\
Heartbeat (405) & 0.697 (0.020) & 0.707 (0.026) & 0.716 (0.014) & 0.723 (0.015) \\
MotorImagery (3000) & 0.534 (0.065) & 0.532 (0.061) & 0.502 (0.019) & 0.524 (0.026) \\
SelfRegulationSCP1 (896) & 0.863 (0.028) & 0.863 (0.012) & 0.884 (0.018) & 0.879 (0.016) \\
SelfRegulationSCP2 (1152) & 0.532 (0.046) & 0.536 (0.044) & 0.534 (0.026) & 0.487 (0.024) \\
StandWalkJump (2500) & 0.413 (0.087) & 0.427 (0.076) & 0.427 (0.076) & 0.440 (0.101) \\
UWaveGestureLibrary (315) & 0.930 (0.009) & 0.949 (0.004) & 0.934 (0.007) & 0.924 (0.006) \\
\end{longtable}

\begin{longtable}{p{4.8cm}cccc}
\caption{Complete per-dataset UEA benchmark results for CNN.}\label{tab:complete_results_uea_cnn}\\
\toprule
Dataset & S=1 & S=2 & S=3 & S=4 \\
\midrule
\endfirsthead
\caption[]{UEA complete benchmark results for CNN (continued)}\\
\toprule
Dataset & S=1 & S=2 & S=3 & S=4 \\
\midrule
\endhead
\midrule
\multicolumn{5}{r}{Continued on next page}\\
\endfoot
\bottomrule
\endlastfoot
AsphObstCoord (736) & 0.633 (0.017) & 0.604 (0.026) & 0.477 (0.109) & 0.275 (0.010) \\
AsphPaveTypeCoord (2371) & 0.752 (0.212) & 0.840 (0.060) & 0.825 (0.022) & 0.851 (0.070) \\
AsphaltRegularityCoordinates (4201) & 0.988 (0.003) & 0.989 (0.003) & 0.991 (0.002) & 0.987 (0.004) \\
AtrialFibrillation (640) & 0.347 (0.119) & 0.320 (0.119) & 0.320 (0.056) & 0.293 (0.076) \\
Cricket (1197) & 0.739 (0.054) & 0.764 (0.033) & 0.822 (0.033) & 0.825 (0.025) \\
DuckDuckGeese (270) & 0.376 (0.036) & 0.300 (0.089) & 0.416 (0.046) & 0.292 (0.067) \\
EigenWorms (17984) & 0.421 (0.003) & 0.424 (0.010) & 0.395 (0.068) & 0.431 (0.020) \\
EthanolConcentration (1751) & 0.299 (0.009) & 0.287 (0.021) & 0.290 (0.011) & 0.275 (0.018) \\
HandMovementDirection (400) & 0.522 (0.035) & 0.541 (0.102) & 0.532 (0.134) & 0.538 (0.080) \\
Heartbeat (405) & 0.737 (0.028) & 0.729 (0.011) & 0.731 (0.019) & 0.700 (0.048) \\
MotorImagery (3000) & 0.480 (0.032) & 0.526 (0.051) & 0.490 (0.027) & 0.520 (0.021) \\
SelfRegulationSCP1 (896) & 0.857 (0.024) & 0.849 (0.023) & 0.779 (0.043) & 0.840 (0.032) \\
SelfRegulationSCP2 (1152) & 0.482 (0.017) & 0.521 (0.024) & 0.493 (0.047) & 0.498 (0.031) \\
StandWalkJump (2500) & 0.347 (0.073) & 0.320 (0.056) & 0.320 (0.030) & 0.467 (0.170) \\
UWaveGestureLibrary (315) & 0.756 (0.036) & 0.756 (0.043) & 0.769 (0.031) & 0.751 (0.018) \\
\end{longtable}

\begin{longtable}{p{4.8cm}cccc}
\caption{Complete per-dataset UEA benchmark results for FCN.}\label{tab:complete_results_uea_fcn}\\
\toprule
Dataset & S=1 & S=2 & S=3 & S=4 \\
\midrule
\endfirsthead
\caption[]{UEA complete benchmark results for FCN (continued)}\\
\toprule
Dataset & S=1 & S=2 & S=3 & S=4 \\
\midrule
\endhead
\midrule
\multicolumn{5}{r}{Continued on next page}\\
\endfoot
\bottomrule
\endlastfoot
AsphObstCoord (736) & 0.808 (0.026) & 0.788 (0.037) & 0.739 (0.033) & 0.765 (0.026) \\
AsphPaveTypeCoord (2371) & 0.963 (0.009) & 0.954 (0.004) & 0.948 (0.004) & 0.949 (0.003) \\
AsphaltRegularityCoordinates (4201) & 0.994 (0.002) & 0.991 (0.006) & 0.990 (0.002) & 0.990 (0.004) \\
AtrialFibrillation (640) & 0.307 (0.060) & 0.293 (0.060) & 0.360 (0.037) & 0.320 (0.030) \\
Cricket (1197) & 0.989 (0.006) & 0.983 (0.015) & 0.969 (0.021) & 0.942 (0.023) \\
DuckDuckGeese (270) & 0.552 (0.039) & 0.564 (0.033) & 0.512 (0.036) & 0.528 (0.039) \\
EigenWorms (17984) & 0.656 (0.226) & 0.666 (0.134) & 0.626 (0.051) & 0.528 (0.032) \\
EthanolConcentration (1751) & 0.294 (0.014) & 0.300 (0.041) & 0.310 (0.019) & 0.297 (0.016) \\
HandMovementDirection (400) & 0.308 (0.055) & 0.400 (0.053) & 0.495 (0.031) & 0.484 (0.037) \\
Heartbeat (405) & 0.686 (0.114) & 0.742 (0.042) & 0.724 (0.037) & 0.670 (0.046) \\
MotorImagery (3000) & 0.506 (0.023) & 0.538 (0.058) & 0.536 (0.030) & 0.534 (0.054) \\
SelfRegulationSCP1 (896) & 0.798 (0.026) & 0.808 (0.042) & 0.837 (0.018) & 0.869 (0.043) \\
SelfRegulationSCP2 (1152) & 0.502 (0.028) & 0.516 (0.050) & 0.510 (0.020) & 0.466 (0.045) \\
StandWalkJump (2500) & 0.347 (0.030) & 0.360 (0.101) & 0.320 (0.056) & 0.280 (0.056) \\
UWaveGestureLibrary (315) & 0.804 (0.013) & 0.909 (0.014) & 0.883 (0.016) & 0.859 (0.016) \\
\end{longtable}
\endgroup

\end{document}